\documentclass[conference]{IEEEtran}
\usepackage{pifont}
\usepackage{times}
\let\labelindent\relax
\usepackage{enumitem}

\usepackage{amsmath}
\usepackage{amssymb}
\usepackage{algorithm}
\usepackage{algpseudocode}
\usepackage{graphicx}
\usepackage{physics,bm,bbm}

\usepackage{multirow}
\usepackage{cuted}
\usepackage{cite}

\usepackage{booktabs}
\usepackage{colortbl}
\usepackage{multicol}
\usepackage{xcolor}
\usepackage{makecell}
\usepackage{subcaption}
\usepackage[colorlinks=true, urlcolor=blue]{hyperref}
\usepackage[numbers]{natbib}
\usepackage{tabularx}
\usepackage{makecell}

\newcommand{\cmark}{\textcolor{green}{\ding{51}}}  
\newcommand{\xmark}{\textcolor{red}{\ding{55}}}    

\begin{document}

\title{AMO: \textbf{A}daptive \textbf{M}otion \textbf{O}ptimization for \\Hyper-Dexterous Humanoid Whole-Body Control}

\author{Jialong Li\textsuperscript{*} $\quad$
        Xuxin Cheng\textsuperscript{*} $\quad$
        Tianshu Huang\textsuperscript{*} $\quad$
        Shiqi Yang $\quad$
        Ri-Zhao Qiu $\quad$
        Xiaolong Wang\\
        UC San Diego \\
        \url{https://amo-humanoid.github.io/}
        \vspace{-0.2in}}

\twocolumn[{%
\renewcommand\twocolumn[1][]{#1}%
\maketitle
\begin{center}
    \vspace{-0.1in}
    \centering
    \captionsetup{type=figure}
    \includegraphics[width=\linewidth]{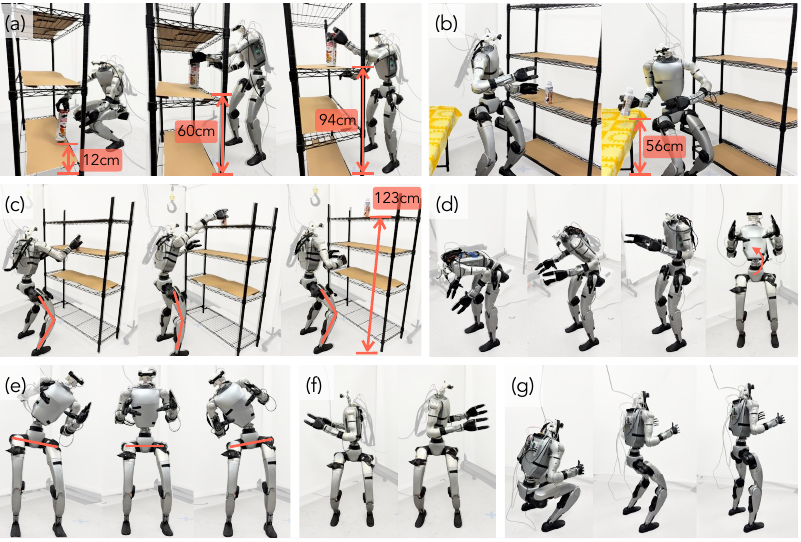}
    \caption{AMO enables hyper-dexterous whole-body movements for humanoid robots. (a): The robot picks and places a can on platforms of different heights. (b): The robot picks a bottle from the higher shelf on the left and puts it on the lower table on the right. (c): The robot stretches its legs to put the bottle on a high shelf. (d-g): The robot demonstrates a wide range of torso pitch, roll, yaw, and height adjustments. (e): The robot utilizes hip motors to compensate waist joint limits to achieve larger roll rotation. }
    \label{fig:teaser}
    \vspace{-0.1in}
\end{center}
}]

\setcounter{footnote}{0}
\renewcommand{\thefootnote}{\fnsymbol{footnote}}
\footnotetext[1]{~Equal contributions. Special thanks to Jiajian Fu for designing and manufacturing of the Active Head 2.0.}
\renewcommand{\thefootnote}{\arabic{footnote}}

\begin{abstract}
Humanoid robots derive much of their dexterity from hyper-dexterous whole-body movements, enabling tasks that require a large operational workspace—such as picking objects off the ground. However, achieving these capabilities on real humanoids remains challenging due to their high degrees of freedom (DoF) and nonlinear dynamics. We propose Adaptive Motion Optimization (AMO), a framework that integrates sim-to-real reinforcement learning (RL) with trajectory optimization for real-time, adaptive whole-body control. To mitigate distribution bias in motion imitation RL, we construct a hybrid AMO dataset and train a network capable of robust, on-demand adaptation to potentially O.O.D. commands. We validate AMO in simulation and on a 29-DoF Unitree G1 humanoid robot, demonstrating superior stability and an expanded workspace compared to strong baselines. Finally, we show that AMO’s consistent performance supports autonomous task execution via imitation learning, underscoring the system’s versatility and robustness.

\end{abstract}

\section{Introduction}
Humans can expand their workspace of hands using whole-body movements. The joint configurations of humanoid robots closely mimic humans' functionality and degree of freedom while facing challenges of achieving similar movements with real-time control. This is due to the dynamic humanoid whole-body control's high-dimensional, highly non-linear, and contact-rich nature. Traditional model-based optimal control methods require precise modeling of the robot and environment, high computational power, and reduced-order models for realizable computation results, which is not feasible for the problem of utilizing all DoFs (29) of an overactuated humanoid robot in the real world. 

Recent advances in reinforcement learning (RL) combined with sim-to-real have demonstrated significant potential in enabling humanoid loco-manipulation tasks in real-world settings~\cite{lu2024pmp}. While these approaches achieve robust real-time control for high-degree-of-freedom (DoF) humanoid robots, they often rely on extensive human expertise and manual tuning of reward functions to ensure stability and performance. To address this limitation, researchers have integrated motion imitation frameworks with RL, leveraging retargeted human motion capture (MoCap) trajectories to define reward objectives that guide policy learning~\cite{he2024omnih2o,cheng2024express}. However, such trajectories are typically kinematically viable but fail to account for the dynamic constraints of target humanoid platforms, introducing an embodiment gap between simulated motions and hardware-executable behaviors. An alternative approach combines trajectory optimization (TO) with RL to bridge this gap~\cite{li2021reinforcement,liu2024opt2skill}.
\begin{table}[t]
    \centering
    \begin{tabularx}{\linewidth}{l
  >{\centering\arraybackslash}X
  >{\centering\arraybackslash}X
  >{\centering\arraybackslash}X
}
    \toprule
     Metrics & \makecell{AMO\\(Ours)} & \makecell{HOVER\\\cite{he2024hover}} & \makecell{Opt2Skill\\ \cite{liu2024opt2skill}} \\
    \midrule
    Ref. Type    &Hybrid &MoCap & Traj. Opt.  \\
    $\mathrm{SO}(3)$ + Height Dex. Torso &\cmark & \xmark   & \xmark \\
    $\mathrm{SE}(3)$ Task Space &\cmark &\xmark   & \xmark \\
    No Ref. Deploy       &\cmark &\cmark & \xmark \\
    O.O.D.    &\cmark &\xmark & \xmark  \\
    \bottomrule
    \end{tabularx}
    \caption{Comparisons with two recent representative humanoid motion imitation works. \textit{Dex. Torso} means if the robot is able to adjust its' torso's orientation and height to expand the workspace. \textit{Task Space} means the end effector. \textit{No Ref. Deploy} means if the robot needs reference motion during deployment. \textit{O.O.D.} means if the work has evaluated O.O.D. performance, which is a typical case when the robot is controlled by a human operator and the control signal is highly unpredictable.}
    \label{tab:comparisons}
    \vspace{-0.1in}
\end{table}

While these methodologies advance humanoid loco-manipulation capabilities, current approaches remain constrained to simplified locomotion patterns rather than achieving true whole-body dexterity. Motion capture-driven methods suffer from inherent kinematic bias: their reference datasets predominantly feature bipedal locomotion sequences (e.g., walking, turning) while lacking coordinated arm-torso movements essential for hyper-dexterous manipulation. Conversely, trajectory optimization TO-based techniques face complementary limitations—their reliance on a limited repertoire of motion primitives and computational inefficiency in real-time applications precludes policy generalization. This critically hinders deployment in dynamic scenarios requiring rapid adaptation to unstructured inputs, such as reactive teleoperation or environmental perturbations.

To bridge this gap, we present Adaptive Motion Optimization (AMO)—a hierarchical framework for real-time whole-body control of humanoid robots through two synergistic innovations: (i) Hybrid Motion Synthesis: We formulate hybrid upper-body command sets by fusing arm trajectories from motion capture data with probabilistically sampled torso orientations, systematically eliminating kinematic bias in the training distribution. These commands drive a dynamics-aware trajectory optimizer to produce whole-body reference motions that satisfy both kinematic feasibility and dynamical constraints, thereby constructing the AMO dataset—the first humanoid motion repository explicitly designed for dexterous loco-manipulation.
(ii) Generalizable Policy Training: While a straightforward solution would map commands to motions via discrete look-up tables, such methods remain fundamentally constrained to discrete, in-distribution scenarios. Our AMO network instead learns continuous mappings, enabling robust interpolation across both continuous input spaces and out-of-distribution (O.O.D.) teleoperation commands while maintaining real-time responsiveness.

During deployment we first extract the sparse poses from a VR teleoperation system and output upper-body goals with multi-target inverse kinematics. The trained AMO network and RL policy together output the robot's control signal. We list a brief comparison in Table. \ref{tab:comparisons} to show the major advantages of our method with two recent representative works. To summarize, our contributions are as follows:
\begin{itemize}
    \item A novel adaptive control method AMO that substantially expands the workspace of humanoid robots. 
    AMO works in real-time with sparse task-space targets and shows O.O.D. performance previous methods have not achieved. 
    \item A new landmark of humanoid hyper-dexterous WBC controller with orders larger workspace that enables a humanoid robot to pick up objects from the ground.
    \item Comprehensive experiments both in simulation and the real world with teleoperation and autonomous results showing the effectiveness of our method and ablations of core components. 
\end{itemize}

\begin{figure*}[t]
    \centering
    \includegraphics[width=\linewidth]{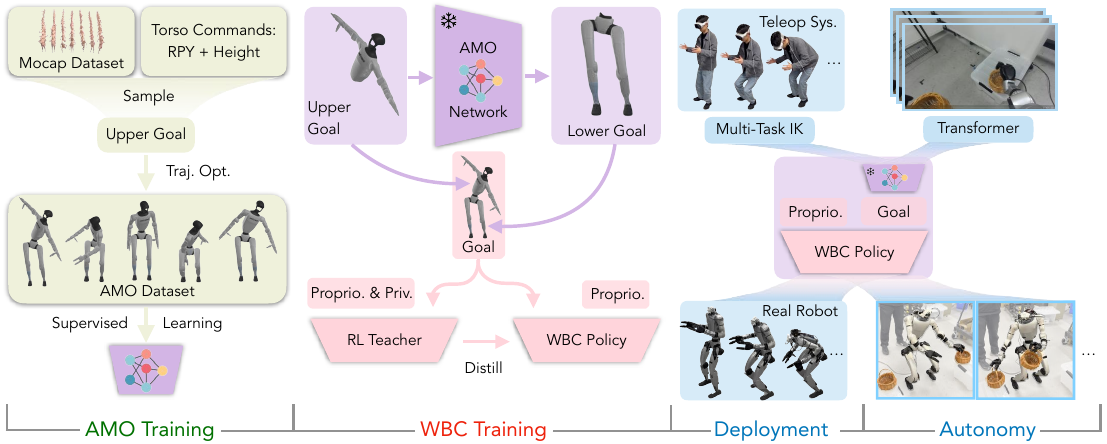}
    \caption{System overview. The system is decomposed into four stages: 1. AMO module training by collecting AMO dataset using trajectory optimization; 2. RL policy training by teacher-student distillation in simulation; 3. real robot teleoperation by IK and retargeting; 4. real robot autonomous policy training by imitation learning (IL) with a transformer.}
    \label{fig:main}
\end{figure*}

\section{Related Work}
\textbf{Humanoid Whole-body Control.}  Whole-body control for humanoid robots remains a challenging problem due to their high DoFs and non-linearity. This is previously primarily achieved by dynamic modeling and model-based control~\cite{miura1984dynamic, yin2007simbicon, hutter2016anymal, moro2019whole, dariush2008whole, kajita2001simple, westervelt2003hybrid, kato1973development, hirai1998development, chignoli2021humanoid, dallard2023Sync, darvish2019wholebody, penco2019multimode}. 
More recently, deep reinforcement learning methods have shown promise in achieving robust locomotion performance for legged robots~\cite{fu2022learning, margolis2022rapid, kumar2021rma, fu2021minimizing, escontrela2022adversarial, li2021reinforcement, siekmann2021blind, agarwal2023legged, yang2023neural, margolis2021learning, duan2023learning, zhuang2023robot, cheng2023extremeparkour, fuchioka2023opt, yang2023generalized, li2024reinforcement, liao2024berkeley, radosavovic2023learning, cheng2023legmanip}.
Researchers have studied whole-body control from high-dimensional inputs for quadrupeds~\cite{fu2022learning, cheng2023extremeparkour, cheng2023legmanip, ha2024umi} and humanoids~\cite{fu2024humanplus, cheng2024exbody, he2024h20, ji2024exbody2}. ~\cite{fu2024humanplus} trains one transformer for control and another transformer for imitation learning. ~\cite{cheng2024exbody} only encourages the upper body to imitate the motion, while the lower body control is decoupled. ~\cite{he2024h20} trains goal-conditioned policies for downstream tasks. All ~\cite{fu2024humanplus, cheng2024exbody, he2024h20} show only limited whole-body control ability, which enforcing the torso and pelvis of humanoid robots to stay motionless. ~\cite{ji2024exbody2} shows expressive whole-body action for humanoid robots, but it does not emphasize leveraging whole-body control to extend robots' loco-manipulation task space.

\textbf{Teleoperation of Humanoids.} Teleoperation of humanoids is crucial for real-time control and robot data collection. Prior efforts in humanoid teleoperation include ~\cite{seo2023deep, he2024h20, he2024omnih2o, fu2024humanplus, cheng2024tv, lu2024pmp}. For example, ~\cite{he2024h20, fu2024humanplus} uses a third-person RGB camera to obtain key points of the human teleoperator. Some works use VR to provide ego-centric observation for teleoperators. ~\cite{cheng2024tv} uses Apple VisionPro to control the active head and upper body with dexterous hands. ~\cite{lu2024pmp} uses Vision Pro for the head and upper body while using pedals for locomotion control. Humanoid whole-body control requires a teleoperator to provide physically achievable whole-body coordinates for the robots.

\textbf{Loco-manipulation Imitation Learning.} Imitation learning has been studied to help the robot complete the task autonomously. Identifying existing work with demonstration sources can be classified as learning from real robot expert data~\cite{brohan2022rt, fu2024mobile, zhao2023learning, shi2024yell, brohan2023rt, khazatsky2024droid, chi2024universal, chi2023diffusionpolicy, pari2021surprising, padalkar2023open, shridhar2023perceiver, Ze2024DP3}, learning from play data~\cite{cui2022play, wang2023mimicplay, mendonca2023alan}, and learning from human demonstrations~\cite{escontrela2022adversarial, Peng2020-ty, fuchioka2023opt, yang2023generalized, peng2021amp, Peng2021-sq, tessler2023calm, wang2020unicon, cheng2024exbody, he2024h20, fu2024humanplus}. Those imitation learning studies are limited to manipulation skills, while there are very few imitation learning studies for loco-manipulation. ~\cite{fu2024mobile} studies imitation learning for loco-manipulation, using a wheel-based robot. This paper uses imitation learning to enable humanoid robots to complete loco-manipulation tasks autonomously.

\section{Adaptive Motion Optimization}

We present AMO: Adaptive Motion Optimization, a framework that achieves seamless whole-body control as illustrated in Figure \ref{fig:main}. Our system is decomposed into four different components. We introduce the notations and overall framework at first, and then elaborate on the learning and realization of these components in the following sections separately.
\subsection{Problem Formulation and Notations}
We address the problem of humanoid whole-body control and focus on two distinct settings: \textbf{teleoperation} and \textbf{autonomous}.

In the teleoperation setting, the whole-body control problem is formulated as learning a goal-conditioned policy $\pi': \mathcal{G} \times \mathcal{S} \rightarrow \mathcal{A}$, where $\mathcal{G}$ represents the goal space, $\mathcal{S}$ the observation space, and $\mathcal{A}$ the action space. In the autonomous setting, the learned policy $\pi: \mathcal{S} \rightarrow \mathcal{A}$ generates actions solely based on observations, without human input.

The goal-conditioned teleoperation policy receives a control signal $\mathbf{g} \in \mathcal{G}$ from the teleoperator, where $\mathbf{g} = [ \mathbf{p}_{\mathrm{head}}, \mathbf{p}_{\mathrm{left}}, \mathbf{p}_{\mathrm{right}}, \mathbf{v} ]$. $\mathbf{p}_{\mathrm{head}}, \mathbf{p}_{\mathrm{left}}, \mathbf{p}_{\mathrm{right}}$ represent the operator's head and hand keypoint poses, while $\mathbf{v} = [ v_x, v_y, v_{\mathrm{yaw}} ]$ specifies the base velocity. Observations $\mathbf{s} \in \mathcal{S}$ include visual and proprioceptive data: $\mathbf{s} = [ \mathrm{img}_{\mathrm{left}}, \mathrm{img}_{\mathrm{right}}, \mathbf{s}_{\mathrm{proprio}} ]$. Actions $\mathbf{a} \in \mathcal{A}$ consist of joint angle commands for the upper and lower body: $\mathbf{a} = [ \mathbf{q}_{\mathrm{upper}}, \mathbf{q}_{\mathrm{lower}} ]$.

 \textit{a) Goal-conditioned teleoperation policy:} The goal-conditioned policy adopts a hierarchical design: $\pi' = [ \pi'_{\mathrm{upper}}, \pi'_{\mathrm{lower}} ]$. The upper policy $\pi'_{\mathrm{upper}} ( \mathbf{p}_{\mathrm{head}}, \mathbf{p}_{\mathrm{left}}, \mathbf{p}_{\mathrm{right}} ) = [ \mathbf{q}_{\mathrm{upper}}, \mathbf{g}' ]$ outputs actions for the upper body and an intermediate control signal $\mathbf{g}' = [ \mathbf{rpy}, h ]$, where $\mathbf{rpy}$ command the torso orientation and $h$ commands the base height. The lower policy $\pi'_{\mathrm{lower}} ( \mathbf{v}, \mathbf{g}', \mathbf{s}_{\mathrm{proprio}} ) = \mathbf{q}_{\mathrm{lower}}$ generates actions for the lower body using this intermediate control signal, the velocity commands, and proprioceptive observations.

\textit{b) Autonomous policy:} The autonomous policy $\pi = [ \pi_{\mathrm{upper}}, \pi_{\mathrm{lower}} ]$ shares the same hierarchical design as the teleoperation policy. The lower policy is identical: $\pi_{\mathrm{lower}} = \pi'_{\mathrm{lower}}$, while the upper policy generates actions and intermediate control independently from human input: $\pi_{\mathrm{upper}} ( \mathrm{img}_{\mathrm{left}}, \mathrm{img}_{\mathrm{right}}, \mathbf{s}_{\mathrm{proprio}} ) = [ \mathbf{q}_{\mathrm{upper}}, \mathbf{v}, \mathbf{g}' ]$.

\subsection{Adaptation Module Pre-Training}
In our system specification, the lower policy follows commands in the form of $[v_x, v_y, v_{\mathrm{yaw}}, \mathbf{rpy}, h]$. The locomotion ability of following velocity commands $[v_x, v_y, v_{\mathrm{yaw}}]$ can be easily learned by randomly sampling directed vectors in the simulation environment following the same strategy as ~\cite{cheng2023extremeparkour, cheng2024exbody}. However, it is non-trivial to learn the torso and height tracking skills as they require whole-body coordination. Unlike in the locomotion task, where we can design feet-tracking rewards based on Raibert Heuristic ~\cite{raibert1986legged} to facilitate skill learning, there lacks such a heuristic to guide the robot to complete whole-body control. Some works ~\cite{ji2024exbody2, he2024omnih2o} train such policies by tracking human references. However, their strategy does not build a connection between human poses and whole-body control directives. 

To address this issue, we propose an \textbf{adaptive motion optimization (AMO)} module. The AMO module is represented as $\phi(\mathbf{q}_{\mathrm{upper}}, \mathbf{rpy}, h) = \mathbf{q}^{\mathrm{ref}}_{\mathrm{lower}}$. Upon receiving whole-body control commands $\mathbf{rpy}, h$ from the upper, it converts these commands into joint angle references for all lower-body actuators for the lower policy to track explicitly. To train this adaptive module, first, we collect an AMO Dataset by randomly sampling upper commands and performing model-based trajectory optimization to acquire lower body joint angles. The trajectory optimization can be formulated as a multi-contact optimal control problem (MCOP) with the following cost fuction:

\begin{align*}
\mathcal{L} &= \mathcal{L}_{\mathbf{x}} + \mathcal{L}_{\mathbf{u}} + \mathcal{L}_{\mathrm{CoM}} + \mathcal{L}_{\mathbf{rpy}} + \mathcal{L}_h\\
\mathcal{L}_{\mathbf{x}} &= \| \mathbf{x}_t - \mathbf{x}_{\mathrm{ref}} \|_{\mathbf{Q}_x}^2 \\
\mathcal{L}_{\mathbf{u}} &= \| \mathbf{u}_t \|_{\mathbf{R}}^2 \\
\mathcal{L}_{\mathrm{CoM}} &= \| \mathbf{c}_t - \mathbf{c}_{\mathrm{ref}} \|_{\mathbf{Q}_{\mathrm{CoM}}}^2 \\
\mathcal{L}_{\mathbf{rpy}} &= \| \mathbf{R}_{\mathrm{torso}} - \mathbf{R}_{\mathrm{ref}}(\mathbf{rpy}) \|_{\mathbf{Q}_{\mathrm{torso}}}^2 \\
\mathcal{L}_h &= w_h (h_t - h)^2
\end{align*}

Which includes regularization for both state $\mathbf{x}$ and control $\mathbf{u}$, target tracking terms $\mathcal{L}_{\mathbf{rpy}}$ and $\mathcal{L}_h$, and a center-of-mass (CoM) regularization term that ensures balance when performing whole-body control. Upon collecting dataset, we first randomly select upper body motions from the AMASS dataset ~\cite{AMASS:2019} and sample random torso commands. Then, we perform trajectory optimizations to track torso objectives while maintaining a stable CoM and adhering to wrench cone constraints to generate dynamically feasible reference joint angles. Since we do not take the walking scenario into consideration, both feet of the robot are considered to be making contact with the ground. The references are generated using control-limited feasibility-driven differential dynamic programming (BoxFDDP) through Crocoddyl ~\cite{mastalli20crocoddyl, mastalli2022fddp}. These data are collected to train an AMO module that converts torso commands into reference lower poses. The AMO module is a three-layer multi-layer perceptron (MLP) and is frozen during the later stage of lower policy training.

\begin{figure}[t]
    \centering
    \includegraphics[width=\linewidth]{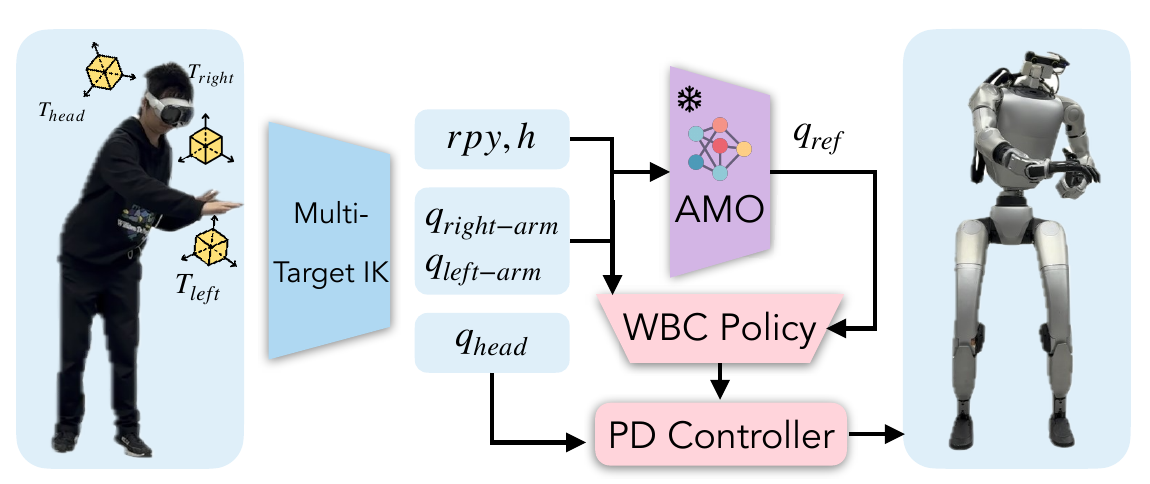}
    \caption{Teleoperation system overview. The operator provides three end-effector targets: head, left wrist, and right wrist poses. A multi-target IK computes upper goals and intermediate goals by matching three weighted targets simultaneously. The intermediate goals ($\mathbf{rpy}, h$) are fed to AMO and converted to lower goals.}
    \label{fig:pipeline}
    \vspace{-0.2in}
\end{figure}

\subsection{Lower Policy Training}
We use massively parallel simulation to train our lower policy with IsaacGym\cite{makoviychuk2021isaac}. The lower policy aims at tracking $\mathbf{g}'$ and $\mathbf{v}$, while utilizing proprioceptive observations $\mathbf{s}_{\mathrm{proprio}}$, which is defined as:

\begin{align*}
[ \bm{\theta}_{t}, \bm{\omega}_{t}, \mathbf{q}^{\mathrm{whole-body}}_{t}, \dot{\mathbf{q}}^{\mathrm{whole-body}}_{t}, \mathbf{a}^{\mathrm{whole-body}}_{t-1}, \bm{\phi}_t, \mathbf{q}^{\mathrm{ref}}_{\mathrm{lower}}]
\end{align*}

The above formulation contains base orientation $\bm{\theta}_{t}$, base angular velocity $\bm{\omega}_{t}$, current position, velocity, and last position targets. It is noticeable that the lower policy's observation includes the states of upper body actuators for better upper-lower body coordination. $\bm{\phi}_t$ is the gait cycling signal defined in similar ways as ~\cite{margolis2022walktheseways, zhang2024whole}. $\mathbf{q}^{\mathrm{ref}}_{\mathrm{lower}}$ is the reference lower body joint angles generated by the AMO module. The lower action space $\mathbf{q}_{\mathrm{lower}} \in \mathbb{R}^{15}$ is a vector of dimension $15$, which consists of $2 * 6$ target joint positions for both legs and $3$ target joint positions for the waist motors. 

We opt to use a teacher-student framework to train our lower policy. We first train a teacher policy that can observe privileged information in simulation, using off-the-shelf PPO\citep{schulman2017proximal}. We then distill the teacher policy into a student policy using supervised learning. The student policy only observes information available in real and can be deployed for teleoperation and autonomous tasks. 

The teacher policy can be formulated as $\pi_{\mathrm{teacher}}(\mathbf{v}, \mathbf{g}', \mathbf{s}_{\mathrm{proprio}}, \mathbf{s}_{\mathrm{priv}}) = \mathbf{q}_{\mathrm{lower}}$. The additional privilege observation $\mathbf{s}_{\mathrm{priv}}$ is defined as:
\begin{align*}
[ \mathbf{v}^{\mathrm{gt}}_{t},  \mathbf{rpy}^{\mathrm{gt}}_t, h^{\mathrm{gt}}_t, \mathbf{c}_t]
\end{align*}

Which includes ground-truth values of: base velocity $\mathbf{v}^{\mathrm{gt}}_{t}$, torso orientation $\mathbf{rpy}^{\mathrm{gt}}_t$, and base height $h^{\mathrm{gt}}_t$, which are not readily available when tracking their corresponding targets in the real world. $c_t$ is the contact signal between feet and the ground. The teacher RL training process is detailed in Appendix.\ref{supp:RL}.

The student policy can be written as $\pi_{\mathrm{student}}(\mathbf{v}, \mathbf{g}', \mathbf{s}_{\mathrm{proprio}}, \mathbf{s}_{\mathrm{hist}}) = \mathbf{q}_{\mathrm{lower}}$. To compensate for $\mathbf{s}_{\mathrm{proprio}}$ with observations accessible in the real-world, the student policy utilizes a 25-step history of proprioceptive observations as an additional input signal: $\mathbf{s}_{\mathrm{hist}, t} =  \mathbf{s}_{\mathrm{proprio}, t-1 \sim t-25}$.

\subsection{Teleoperation Upper Policy Implementation}
The teleoperation upper policy generates a series of commands for whole-body control, including arm and hand movements, torso orientation, and base height. 
We opt to implement this policy using optimization-based techniques to achieve the precision required for manipulation tasks. Specifically, hand movements are generated via retargeting, while other control signals are computed using inverse kinematics (IK). Our implementation of hand retargeting is based on dex-retargeting~\cite{qin2023anyteleop}. More details about our retargeting formulation are presented in Appendix.\ref{supp:retargeting}. 

In our whole body control framework, we extend the conventional IK to multi-target weighted IK, minimizing the 6D distances to three key targets: the head, the left wrist, and the right wrist. The robot mobilizes all upper-body actuators to match these three targets simultaneously. Formally, our objective is:

\begin{align*}
\min_{\mathbf{q}} &\quad \mathcal{L}_\mathrm{head} + \mathcal{L}_\mathrm{left} + \mathcal{L}_\mathrm{right}\\
\mathcal{L}_\mathrm{head} &= \| \mathbf{p}_\mathrm{head} - \mathbf{p}_\mathrm{head-link} \|^2 + \lambda \| \mathbf{R}_\mathrm{head} - \mathbf{R}_\mathrm{head-link} \|_F^2\\
\mathcal{L}_\mathrm{left} &= \| \mathbf{p}_\mathrm{left} - \mathbf{p}_\mathrm{left-link} \|^2 + \lambda \| \mathbf{R}_\mathrm{left} - \mathbf{R}_\mathrm{left-link} \|_F^2\\
\mathcal{L}_\mathrm{right} &= \| \mathbf{p}_\mathrm{right} - \mathbf{p}_\mathrm{right-link} \|^2 + \lambda \| \mathbf{R}_\mathrm{right} - \mathbf{R}_\mathrm{right-link} \|_F^2\\
\mathbf{q} &= [\mathbf{q}_\mathrm{head}, \mathbf{q}_\mathrm{left-arm}, \mathbf{q}_\mathrm{right-arm}, \mathbf{rpy}, h]
\end{align*}

As illustrated in Fig. \ref{fig:pipeline}. The optimization variable $\mathbf{q}$ includes all actuated degrees of freedom (DoFs) in the robot's upper body: $\mathbf{q}_\mathrm{head}$, $\mathbf{q}_\mathrm{left-arm}$, and $\mathbf{q}_\mathrm{right-arm}$. In addition to the motor commands, it also solves for an intermediate command to enable whole-body coordination: $\mathbf{rpy}$ and $h$ for torso orientation and height control. To ensure smooth upper body control, posture costs are weighted differently across different components of the optimization variable $\mathbf{q}$: $\mathbf{W}_{\mathbf{q}_\mathrm{head}, \mathbf{q}_\mathrm{left-arm}, \mathbf{q}_\mathrm{right-arm}} < \mathbf{W}_{\mathbf{rpy}, h}$. This encourages the policy to prioritize upper-body actuators for simpler tasks. However, for tasks requiring whole-body movement, such as bending over to pick up or reaching distant targets, additional control signals $[\mathbf{rpy}, h]$ are generated and sent to the lower policy. The lower policy coordinates its motor angles to fulfill the upper policy's requirements, enabling whole-body target reaching. Our IK implementation employs Levenberg-Marquardt (LM) algorithm ~\cite{Sugihara2011} and is based on Pink~\cite{pink2024}.

\subsection{Autonomous Upper Policy Training}
We learn the autonomous upper policy via imitation learning. First, the human operators teleoperate with the robot using our goal-conditioned policy, recording observations and actions as demonstrations. We then employ ACT~\cite{zhao2023learning} with a DinoV2~\cite{oquab2023dinov2, darcet2023vitneedreg} visual encoder as the policy backbone. The visual observation includes two stereo images $\mathrm{img}_{\mathrm{left}}$ and $\mathrm{img}_{\mathrm{right}}$. DinoV2 divides each image into $16 \cross 22$ patches and produces a 384-dimensional visual token for each patch, yielding a combined visual token of shape $2 \cross 16 \cross 22 \cross 384$. This visual token is concatenated with a state token obtained by projecting $\mathbf{o}_t = [\mathbf{s}^\mathrm{upper}_{\mathrm{proprio}, t}, \mathbf{v}_{t-1}, \mathbf{rpy}_{t-1}, h_{t-1}]$. Here, $\mathbf{s}^\mathrm{upper}_{\mathrm{proprio}, t}$ is the upper-body proprioceptive observation, and $[\mathbf{v}_{t-1}, \mathbf{rpy}_{t-1}, h_{t-1}]$ constitutes the last command sent to the lower policy. Due to our decoupled system design, the upper policy observes these lower-policy commands instead of direct lower-body proprioception. The policy's output is represented as:

\begin{align*}
[ \mathbf{q}^{\mathrm{head}}_{t}, \mathbf{q}^{\mathrm{dual-arm}}_{t}, \mathbf{q}^{\mathrm{dual-hand}}_{t}, \mathbf{v}_{t}, \mathbf{rpy}_{t}, h_{t} ]
\end{align*}

comprising all upper-body joint angles and the intermediate control signals for the lower policy.

\begin{table}[t]
    \centering
    \resizebox{\linewidth}{!}{
    \begin{tabular}{@{}lccccc@{}}
        \toprule
        Metrics ($\downarrow$) & $E_y$ & $E_p $ & $E_r$ & $E_h$ & $E_{\mathbf{v}}$ \\ 
        \midrule
        \rowcolor{lightgray}
        \multicolumn{6}{l}{\textbf{Stand}} \\
        \midrule
        Ours(AMO) & 0.1355 & \textbf{0.1259} & \textbf{0.0675} & \textbf{0.0151} & \textbf{0.0465}\\ 
        w/o AMO & 0.3137 & 0.1875 & 0.0681 & 0.1168 & 0.0477 \\ 
        w/o priv & 0.1178 & 0.1841 & 0.0757 & 0.0228 & 0.1771 \\ 
        w rand arms & \textbf{0.1029} & 0.1662 & 0.0716 & 0.0200 & 0.1392 \\
        \midrule
        \rowcolor{lightgray}
        \multicolumn{6}{l}{\textbf{Walk}} \\
        \midrule
        Ours(AMO) & 0.1540 & \textbf{0.1519} & \textbf{0.0735} & \textbf{0.0182} & 0.1779 \\ 
        w/o AMO & 0.3200 & 0.1927 & 0.0797 & 0.1253 & \textbf{0.1539} \\ 
        w/o priv & 0.1226 & 0.1879 & 0.0779 & 0.0276 & 0.2616 \\ 
        w rand arms & \textbf{0.1200} & 0.1837 & 0.0790 & 0.0240 & 0.2596 \\
        \bottomrule
    \end{tabular}
    }
    \caption{Comparison of tracking errors with baselines. Each tracking error is averaged over 4096 environments and 500 steps. $E_y$, $E_p$, $E_r$, $E_h$, $E_{\mathbf{v}}$ represent tracking errors in torso yaw, torso pitch, torso roll, base height, and base linear velocity, respectively.}
    \label{tab:errors_comparison}
    \vspace{-0.2in}
\end{table}

\section{Evaluation}

In this section, we aim to address the following questions by conducting experiments in both simulation and in the real world:

\begin{itemize}
\item How well does AMO performs on tracking locomotion commands $\mathbf{v}$ and torso commands $\mathbf{rpy}, h$?
\item How does AMO compare to other WBC strategies?
\item How well does the AMO system perform in the real-world setting?
\end{itemize}

We conduct our sim experiments in IsaacGym simulator ~\cite{makoviychuk2021isaac}. Our real robot setup is as shown in \ref{fig:pipeline}, which is modified from Unitree G1 ~\cite{unitree2024g1} with two Dex3-1 dexterous hands. This platform features 29 whole-body DoFs and 7 DoFs on each hand. We customized an active head with three actuated DoFs to map the human operator's head movement and mounted a ZED Mini~\cite{zed} camera for stereo streaming.

\subsection{How well does AMO perform on tracking locomotion commands $\mathbf{v}$ and torso commands $\mathbf{rpy}, h$?} Table ~\ref{tab:errors_comparison} presents the evaluation of AMO's performance by comparing it against the following baselines: 
\begin{itemize}
\item \textbf{w/o AMO:} This baseline follows the same RL training recipe as \textbf{Ours(AMO)}, with two key modifications. First, it excludes the AMO output $\mathbf{q}^\mathrm{ref}_\mathrm{lower}$ from the observation space. Second, instead of penalizing deviations from $\mathbf{q}^\mathrm{ref}_\mathrm{lower}$, it applies a regularization penalty based on deviations from a default stance pose.
\item \textbf{w/o priv:} This baseline is trained without additional privileged observations $\mathbf{s}_{\mathrm{priv}}$.
\item \textbf{w rand arms:} In this baseline, arm joint angles are not set using human references sampled from a MoCap dataset. Instead, they are randomly assigned by uniformly sampling values within their respective joint limits.
\end{itemize}

\begin{table}[t!]
    \centering
    \begin{tabular}{@{}lccc@{}}
        \toprule
        Metrics & $R_y$ & $R_p$ & $R_r$\\ 
        \midrule
        Ours(AMO) & (\textbf{-1.5512}, \textbf{1.5868}) & (-0.4546, \textbf{1.5745}) & (\textbf{-0.4723}, \textbf{0.4714}) \\ 
        w/o AMO & (-0.9074, 0.8956) & (\textbf{-0.4721}, 1.0871) & (-0.3921, 0.3626) \\ 
        Waist Tracking & (-1.5274, 1.5745) & (-0.4531, 0.5200) & (-0.3571, 0.3068) \\ 
        ExBody2 & (-0.1005, 0.0056) & (-0.2123, 0.4637) & (-0.0673, 0.0623) \\ 
        \bottomrule
    \end{tabular}
    \caption{Comparison of maximum torso control ranges. $R_y, R_p, R_r$ is the maximum range the torso can achieve while following in-distribution (I.D.) tracking commands in yaw, pitch, roll directions.}
    \label{tab:torso_ranges_comparison}
    \vspace{-0.2in}
\end{table}

The performance is evaluated using the following metrics:

\begin{enumerate}
\item \textbf{Torso Orientation Tracking Accuracy:} Torso orientation tracking is measured by $E_y$, $E_p$, $E_r$. The results indicate that \textit{AMO} achieves superior tracking accuracy in roll and pitch directions. The most notable improvement is in pitch tracking, where other baselines struggle to maintain accuracy, whereas our model significantly reduces tracking error. \textit{w rand arms} exhibits the lowest yaw tracking error, potentially because random arm movements enable the robot to explore a broader range of postures. However, \textit{AMO} is not necessarily expected to excel in yaw tracking, as torso yaw rotation induces minimal CoM displacement compared to roll and pitch. Consequently, yaw tracking accuracy may not fully capture AMO's capacity for generating adaptive and stable poses. Nonetheless, it is worth noting that \textit{w/o AMO} struggles with yaw tracking, suggesting that AMO provides critical reference information for achieving stable yaw control.
\item \textbf{Height Tracking Accuracy:} The results show that \textit{AMO} achieves the lowest height tracking error. Notably, \textit{w/o AMO} reports a significantly higher error than all other baselines, indicating that it barely tracks height command. Unlike torso tracking, where at least one waist motor angle is directly proportional to the command, height tracking requires coordinated adjustments across multiple lower-body joints. Without reference information from AMO, the policy fails to learn the transformation relationship between height commands and corresponding motor angles, making it difficult to master this skill.
\item \textbf{Linear Velocity Tracking Accuracy:} The AMO module generates reference poses based on whole-body control in a double-support stance, meaning it does not account for pose variations due to foot swing during locomotion. Despite this limitation, \textit{AMO} remains capable of performing stable locomotion with a low tracking error, demonstrating its robustness. 
\end{enumerate}

\subsection{How does AMO compare to other WBC strategies?}

\begin{figure}[t]
    \centering
    \includegraphics[width=\linewidth]{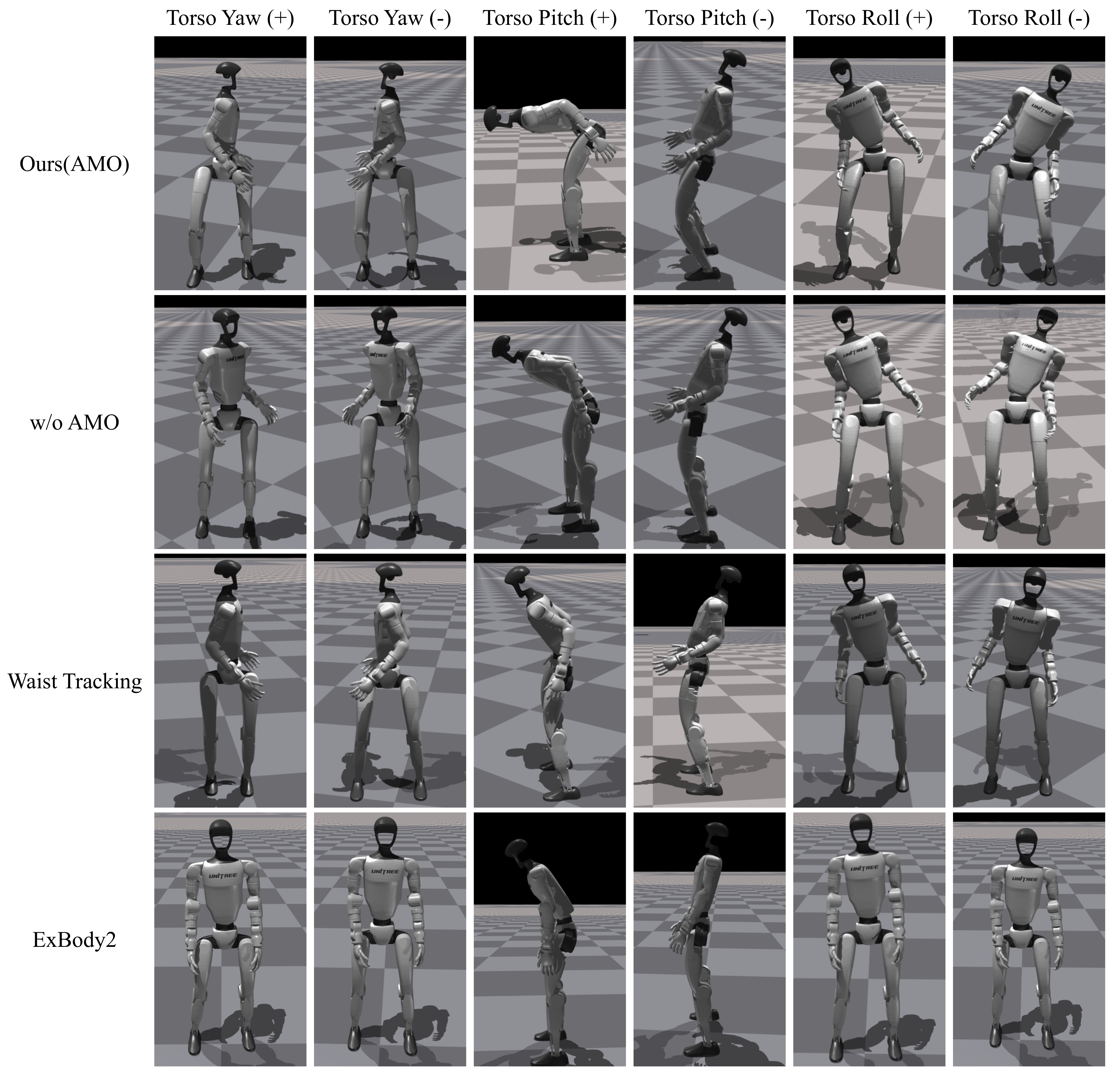}
    \caption{Comparison of torso orientation ranges.}
    \label{fig:ranges}
    \vspace{-0.2in}
\end{figure}

To address this question, we compare the range of torso control across the following baselines:
\begin{itemize}
\item \textbf{w/o AMO:} This baseline is the same as described in the previous section.
\item \textbf{Waist Tracking:} This baseline explicitly commands the yaw, roll, and pitch angles of the waist motors instead of controlling the torso orientation.
\item \textbf{ExBody2:} ~\cite{ji2024exbody2} is a representative work in leveraging human reference motions to guide robot whole-body control in RL. It achieves torso orientation control by modifying the reference joint angles of the waist.
\end{itemize}

\begin{figure}[t]
    \centering
    \includegraphics[width=\linewidth]{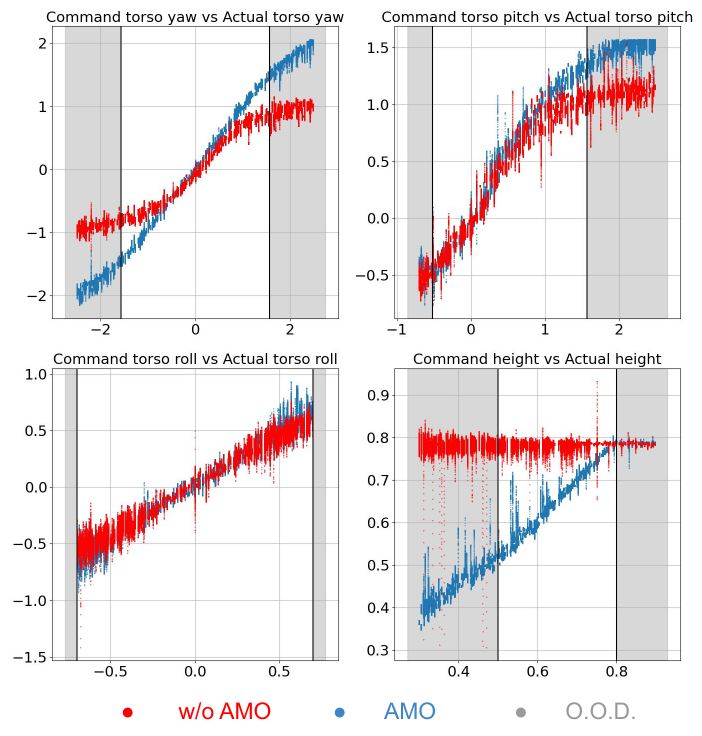}
    \caption{Evaluation of in-distribution (I.D.) and out-of-distribution (O.O.D.) tracking results. Each figure shows the target vs. the actual commanded direction. The white area indicates I.D., meaning the command is used in both trajectory optimization and RL training. The grey area indicates O.O.D., meaning the command is not used in either trajectory optimization or RL training. The red and blue curves represent \textit{w.o. AMO} and \textit{AMO}, respectively.}
    \label{fig:ood}
    \vspace{-0.2in}
\end{figure}

Table.~\ref{tab:torso_ranges_comparison} presents the quantitative measurements of torso control ranges, while Fig.~\ref{fig:ranges} illustrates the qualitative differences. For \textit{ExBody2} and other methods that rely on human motion tracking, the range of torso control is inherently constrained by the diversity of the human motion dataset used for training. If a particular movement direction is underrepresented in the dataset, the learned policy struggles to generalize beyond those sparse examples. As shown in our results, \textit{ExBody2} exhibits only minor control over torso pitch and is largely incapable of tracking torso yaw and roll movements. 

\textit{Waist Tracking} is fundamentally restricted by the robot's waist joint limits, as it relies exclusively on waist motors for torso orientation control rather than utilizing the entire lower body. For example, Unitree G1's waist pitch motor's positional limit is only $\pm 0.52$ radians. In contrast, \textit{AMO} achieves a significantly larger range of torso motion compared to other baselines, particularly in torso pitch, where it allows the robot to bend its upper body completely flat. Moreover, the policy demonstrates adaptive behaviours by leveraging leg motors to adjust the lower posture for stability. This is evident in torso roll control, where the robot slightly bends one leg to tilt its pelvis. Such adaptive behaviour is not observed in \textit{w/o AMO}. Overall, by incorporating AMO, the policy not only expands the operational range of torso control but also improves torso stability by dynamically adapting to the commanded orientation.

\begin{figure}[t]
    \centering
    \begin{subfigure}[t]{\linewidth}
    \centering
    \includegraphics[width=\linewidth]{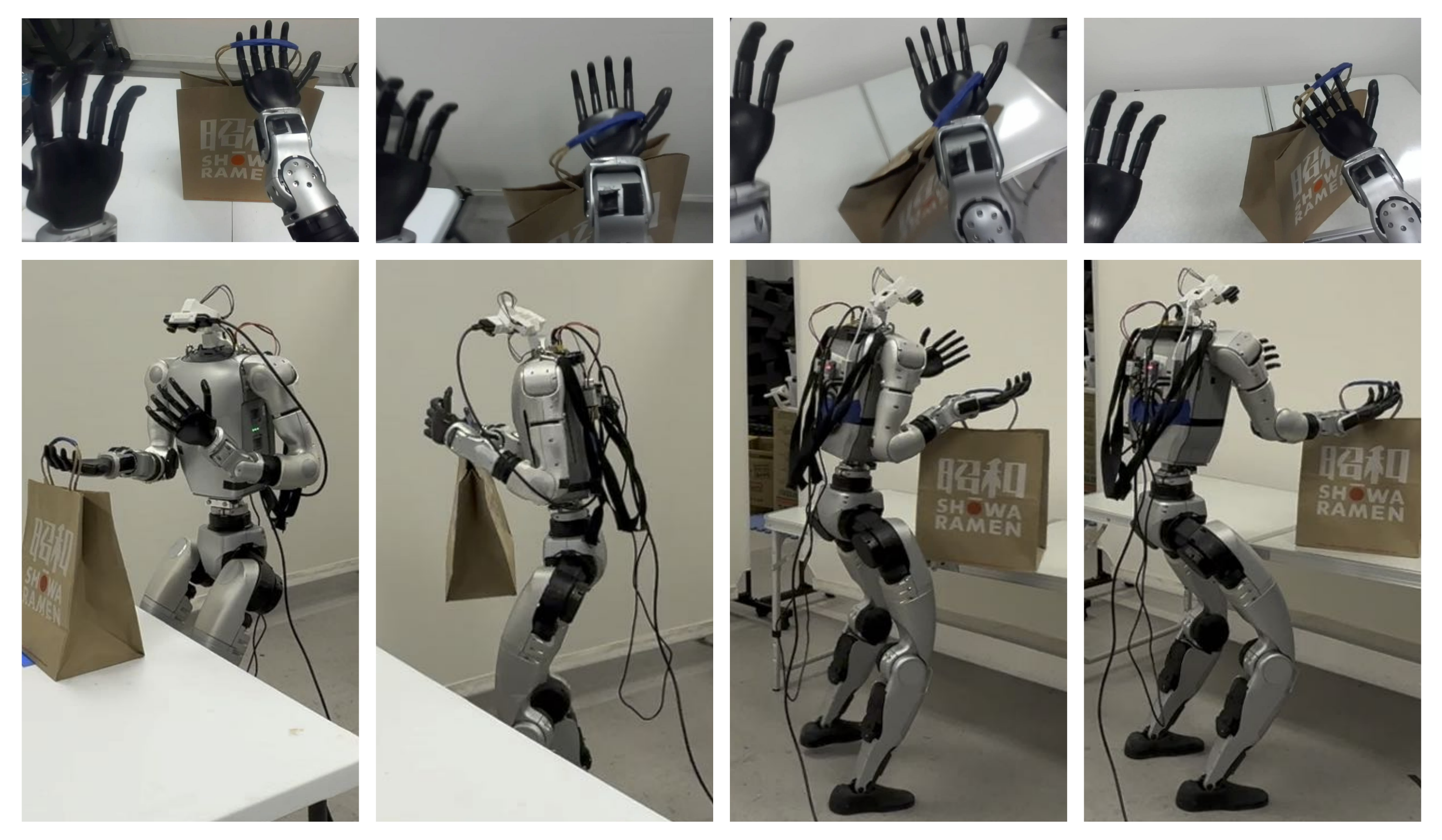}
    \caption{\textbf{Paper Bag Picking:} The task begins with the robot adjusting its torso to align its rubber hand with the handle. Then, the robot should turn and move to an appropriate distance to the destination table, stand still, and use its upper body joints coherently to leave the hand out of the handle.}
    \label{fig:bag-picking}
    \end{subfigure} 
    \begin{subfigure}[t]{\linewidth}
    \centering
    \includegraphics[width=\linewidth]{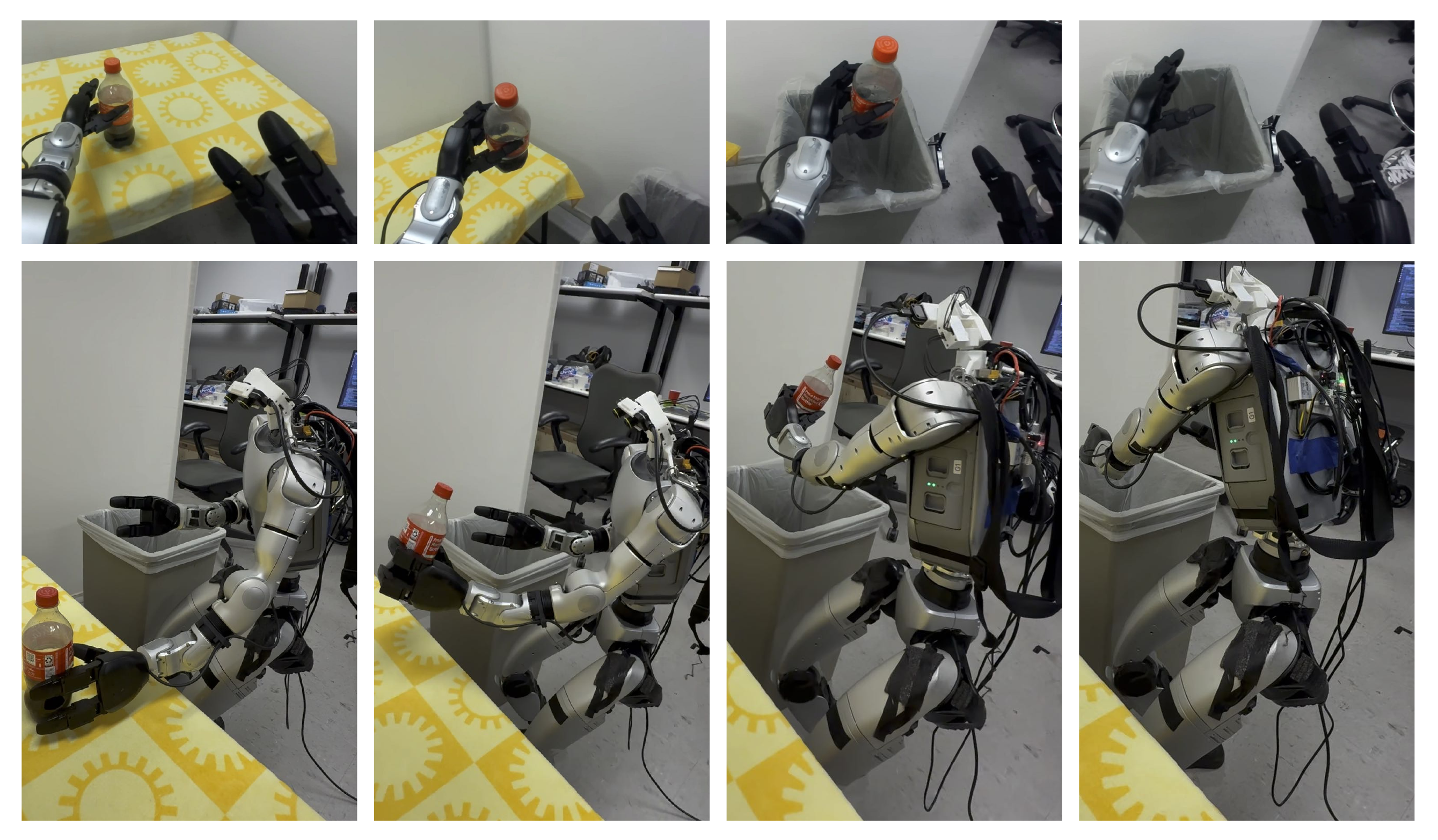}
    \caption{\textbf{Trash Bottle Throwing:} The task begins with the robot stooping and turning its upper body to the left to grab the trash bottle. The robot then turns its waist about 90 degrees to the right and throws the trash bottle into the trash bin.}
    \label{fig:bottle-throwing}
    \end{subfigure}
    \caption{Autonomous tasks performed in the real-world setting. For each task, we collect 50 episodes using the teleoperation system and train an ACT to complete it autonomously.}
    \label{fig:tasks}
    \vspace{-0.2in}
\end{figure}

AMO's advantage lies not only in accurately tracking in-distribution (I.D.) torso commands but also in its ability to effectively adapt to out-of-distribution (O.O.D.) commands. In Fig.\ref{fig:ood}, we compare \textit{AMO} and \textit{w/o AMO} by evaluating their performance on both I.D. and O.O.D. commands. It is evident that \textit{w/o AMO} struggles with O.O.D. commands: it fails to track torso pitch and yaw commands before they reach the sampled training ranges, and it does not track height commands at all, as discussed in the previous section. In contrast, \textit{AMO} exhibits remarkable adaptability in tracking O.O.D. commands. It successfully tracks torso yaw command up to $\pm 2$, despite being trained only within the range of $\pm 1.57$. Similarly, for height tracking, although the training distribution was limited to a range of $0.5m$ to $0.8m$, the policy generalizes well and accurately tracks height as low as $0.4m$. These results indicate that both the AMO module trained via imitation and the RL policy exhibit strong generalization capabilities, demonstrating AMO’s robustness in adapting to whole-body commands beyond the training distribution.

\subsection{How well does the AMO system perform in the real-world setting?}

To showcase our policy's capability in torso orientation and base height control, we have performed a series of hyper-dexterous manipulation tasks using the AMO teleoperation framework. These teleoperation experiments are presented in Fig.\ref{fig:teaser} and the supplementary videos.

To further highlight the robustness and hyper-dexterity of the AMO system, We select several challenging tasks that require adaptive whole-body control and perform imitation learning by collecting demonstrations, as illustrated in Fig.\ref{fig:tasks}. The performance and evaluations of these tasks are detailed below:

\begin{table}[t!]
    \centering
    \begin{tabularx}{\linewidth}{@{}X c c c@{}}
        \toprule
        \rowcolor{lightgray}
        \multicolumn{4}{l}{\textbf{Paper Bag Picking}} \\
        \midrule
        Setting & Picking & Moving & Placing \\ 
        \midrule
        Stereo Input; Chunk Size 120 & 8/10 & 8/10 & 9/10 \\ 
        Mono Input; Chunk Size 120 & 9/10 & 7/10 & 10/10 \\ 
        Stereo Input; Chunk Size 60 & 7/10 & 7/10 & 6/10 \\ 
    \end{tabularx}
    \begin{tabularx}{\linewidth}{@{}X c c@{}}
        \toprule
        \rowcolor{lightgray}
        \multicolumn{3}{l}{\textbf{Trash Bottle Throwing}} \\
        \midrule
        Setting & Picking & Placing \\ 
        \midrule
        Stereo Input; Chunk Size 120 & 7/10 & 10/10 \\ 
        Mono Input; Chunk Size 120 & 4/10 & 10/10 \\ 
        Stereo Input; Chunk Size 60 & 5/10 & 10/10 \\ 
        \bottomrule
    \end{tabularx}
    \caption{Training settings and success rate of individual stages of each task. Each training setting includes: using stereo (both left-right-eye images) or mono (single image) visual input; chunk size used in training action-chunking-transformer.}
    \label{tab:task_rate}
    \vspace{-0.2in}
\end{table}

\textbf{Paper Bag Picking:} This task requires the robot to perform a loco-manipulation task in the absence of an end-effector, making it particularly demanding in terms of precision. We evaluate the impact of various training settings on task performance, as shown in Table.\ref{tab:task_rate}. With the most complete setting, the policy achieves a near-perfect success rate. While using only a single image slightly reduces the success rate, this task is particularly susceptible to shorter chunk sizes. When the robot places its hand under the bag handle and attempts to reorient, a shorter action memory often leads to confusion.
 
\textbf{Trash Bottle Throwing:} This task involves no locomotion; however, to successfully grasp the bottle and throw it into a bin positioned at another angle, the robot must execute extensive torso movements. The evaluation results, as shown in Table.\ref{tab:task_rate}, indicate that our system can learn to complete this task autonomously. Both stereo vision and longer action chunks enhance task performance. The benefits of stereo vision likely stem from an increased field-of-view (FoV) and implicit depth information, which are essential for grasping.

\begin{figure}[t]
    \centering
    \includegraphics[width=\linewidth]{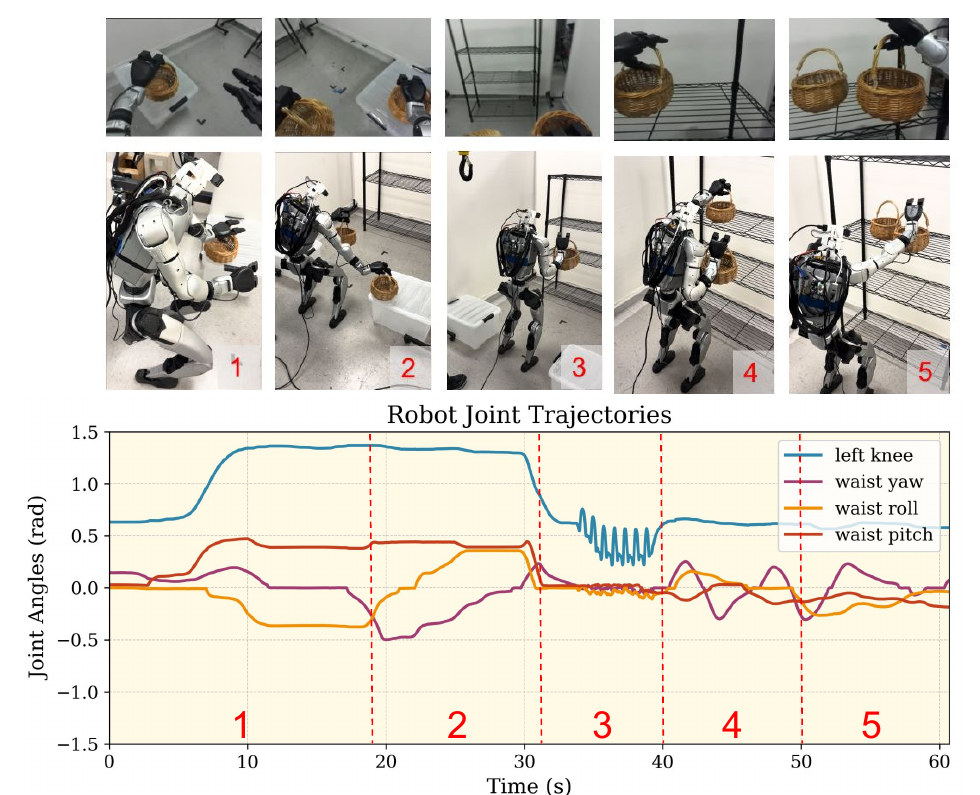}
    \caption{\textbf{Basket Picking:} A complicated loco-manipulation task that also requires whole-body coordination. The task begins with the robot picking two baskets from left(1) and right(2), which are placed at low heights and close to the ground. Then, the robot stands up, moves forward(3), and puts two baskets on the shelf at eye level(4,5). The motor angle trajectories along the autonomous rollout are displayed and labeled to match the corresponding phases.}
    \label{fig:case_study}
    \vspace{-0.2in}
\end{figure}

To demonstrate the effectiveness of our system and its potential in executing complex loco-manipulation tasks, we conduct an IL experiment on a long-horizon task that demands hyper-dexterous whole-body control: \textbf{Basket Picking.} In this task, the robot must crouch to a considerably low height and adjust its torso orientation to grasp two baskets positioned on either side, close to the ground. A recorded autonomous rollout is presented as a case study in Fig.\ref{fig:case_study}, showcasing the learned policy's execution in real-world conditions.

To illustrate how the robot coordinates its whole-body actuators, we visualize the joint angles of the waist motors and the left knee motor. The left knee motor is selected as a representative joint for height changing and walking. As shown in Fig.\ref{fig:case_study}, the task begins with the robot crouching and bending forward to align its hands with the baskets' height. This motion is reflected by an increase in the knee and waist pitch motor angles. Next, the robot tilts left and right to grasp the baskets, as indicated by the variations in the waist roll curve. After successfully retrieving the targets, the robot stands up, which is marked by a decrease in knee angle. The periodic fluctuations in the knee motor reading confirm that the robot is walking. Once stationary, the robot reaches to the left and right to place the basket on the shelf, with the waist yaw motor rotating back and forth to facilitate lateral torso movements. This case study clearly demonstrates our system's ability to leverage whole-body coordination to accomplish complex tasks effectively.

\section{Conclusions and Limitations} 
\label{sec:conclusion}

We present \textbf{AMO}, a framework that integrates model-based trajectory optimization with model-free reinforcement learning to achieve whole-body control. Through extensive experiments in both simulation and real-robot platforms, we demonstrate the effectiveness of our approach. AMO enables real-world humanoid control with an unprecedented level of dexterity and precision. We hope that our framework provides a novel pathway for achieving humanoid whole-body control beyond conventional model-based whole-body methods and RL policies that rely on tracking human motion references.

While our decoupled approach introduces a new paradigm for whole-body control, its separated nature inherently limits the level of whole-body coordination it can achieve. For instance, in highly dynamic scenarios, humans naturally utilize not only their lower body and waist but also their arms to maintain balance. However, in our current setup, arm control is independent of the robot's base state. A balance-aware upper-body control mechanism could be explored by incorporating base state information into upper-limb joint-angle acquisition, potentially enhancing overall stability and adaptability.

\section{Acknowledgements} 
\label{sec:acknowledgement}
This project was supported, in part, by gifts from Amazon, Qualcomm and Meta.

\bibliographystyle{plainnat}
\bibliography{references}

\begin{thebibliography}{79}
\providecommand{\natexlab}[1]{#1}
\providecommand{\url}[1]{\texttt{#1}}
\expandafter\ifx\csname urlstyle\endcsname\relax
  \providecommand{\doi}[1]{doi: #1}\else
  \providecommand{\doi}{doi: \begingroup \urlstyle{rm}\Url}\fi

\bibitem[uni()]{unitree2024g1}
{Unitree} {Robotics}, {G1}, 2024, \url{www.unitree.com/g1}, [Online; accessed Jan. 2025].

\bibitem[zed()]{zed}
{Zed} {Mini}, {Stereo} {Camera}, 2024, \url{www.robotis.us/dynamixel-xl330-m288-t}, [Online, accessed Jun. 2024].

\bibitem[Agarwal et~al.(2023)Agarwal, Kumar, Malik, and Pathak]{agarwal2023legged}
Ananye Agarwal, Ashish Kumar, Jitendra Malik, and Deepak Pathak.
\newblock Legged locomotion in challenging terrains using egocentric vision.
\newblock In \emph{Conference on Robot Learning}, pages 403--415. PMLR, 2023.

\bibitem[Brohan et~al.(2022)Brohan, Brown, Carbajal, Chebotar, Dabis, Finn, Gopalakrishnan, Hausman, Herzog, Hsu, et~al.]{brohan2022rt}
Anthony Brohan, Noah Brown, Justice Carbajal, Yevgen Chebotar, Joseph Dabis, Chelsea Finn, Keerthana Gopalakrishnan, Karol Hausman, Alex Herzog, Jasmine Hsu, et~al.
\newblock Rt-1: Robotics transformer for real-world control at scale.
\newblock \emph{arXiv preprint arXiv:2212.06817}, 2022.

\bibitem[Brohan et~al.(2023)Brohan, Brown, Carbajal, Chebotar, Chen, Choromanski, Ding, Driess, Dubey, Finn, et~al.]{brohan2023rt}
Anthony Brohan, Noah Brown, Justice Carbajal, Yevgen Chebotar, Xi~Chen, Krzysztof Choromanski, Tianli Ding, Danny Driess, Avinava Dubey, Chelsea Finn, et~al.
\newblock Rt-2: Vision-language-action models transfer web knowledge to robotic control.
\newblock \emph{arXiv preprint arXiv:2307.15818}, 2023.

\bibitem[Caron et~al.(2024)Caron, De~Mont-Marin, Budhiraja, Bang, Domrachev, and Nedelchev]{pink2024}
Stéphane Caron, Yann De~Mont-Marin, Rohan Budhiraja, Seung~Hyeon Bang, Ivan Domrachev, and Simeon Nedelchev.
\newblock {Pink: Python inverse kinematics based on Pinocchio}, 2024.
\newblock URL \url{https://github.com/stephane-caron/pink}.

\bibitem[Cheng et~al.(2023{\natexlab{a}})Cheng, Kumar, and Pathak]{cheng2023legmanip}
Xuxin Cheng, Ashish Kumar, and Deepak Pathak.
\newblock Legs as manipulator: Pushing quadrupedal agility beyond locomotion.
\newblock In \emph{2023 IEEE International Conference on Robotics and Automation (ICRA)}, 2023{\natexlab{a}}.

\bibitem[Cheng et~al.(2023{\natexlab{b}})Cheng, Shi, Agarwal, and Pathak]{cheng2023extremeparkour}
Xuxin Cheng, Kexin Shi, Ananye Agarwal, and Deepak Pathak.
\newblock Extreme parkour with legged robots, 2023{\natexlab{b}}.
\newblock URL \url{https://arxiv.org/abs/2309.14341}.

\bibitem[Cheng et~al.(2024{\natexlab{a}})Cheng, Ji, Chen, Yang, Yang, and Wang]{cheng2024exbody}
Xuxin Cheng, Yandong Ji, Junming Chen, Ruihan Yang, Ge~Yang, and Xiaolong Wang.
\newblock Expressive whole-body control for humanoid robots, 2024{\natexlab{a}}.
\newblock URL \url{https://arxiv.org/abs/2402.16796}.

\bibitem[Cheng et~al.(2024{\natexlab{b}})Cheng, Ji, Chen, Yang, Yang, and Wang]{cheng2024express}
Xuxin Cheng, Yandong Ji, Junming Chen, Ruihan Yang, Ge~Yang, and Xiaolong Wang.
\newblock Expressive whole-body control for humanoid robots.
\newblock \emph{arXiv preprint arXiv:2402.16796}, 2024{\natexlab{b}}.

\bibitem[Cheng et~al.(2024{\natexlab{c}})Cheng, Li, Yang, Yang, and Wang]{cheng2024tv}
Xuxin Cheng, Jialong Li, Shiqi Yang, Ge~Yang, and Xiaolong Wang.
\newblock Open-television: Teleoperation with immersive active visual feedback.
\newblock \emph{CoRL}, 2024{\natexlab{c}}.

\bibitem[Chi et~al.(2023)Chi, Feng, Du, Xu, Cousineau, Burchfiel, and Song]{chi2023diffusionpolicy}
Cheng Chi, Siyuan Feng, Yilun Du, Zhenjia Xu, Eric Cousineau, Benjamin Burchfiel, and Shuran Song.
\newblock Diffusion policy: Visuomotor policy learning via action diffusion.
\newblock In \emph{Proceedings of Robotics: Science and Systems (RSS)}, 2023.

\bibitem[Chi et~al.(2024)Chi, Xu, Pan, Cousineau, Burchfiel, Feng, Tedrake, and Song]{chi2024universal}
Cheng Chi, Zhenjia Xu, Chuer Pan, Eric Cousineau, Benjamin Burchfiel, Siyuan Feng, Russ Tedrake, and Shuran Song.
\newblock Universal manipulation interface: In-the-wild robot teaching without in-the-wild robots.
\newblock In \emph{Proceedings of Robotics: Science and Systems (RSS)}, 2024.

\bibitem[Chignoli et~al.(2021)Chignoli, Kim, Stanger-Jones, and Kim]{chignoli2021humanoid}
Matthew Chignoli, Donghyun Kim, Elijah Stanger-Jones, and Sangbae Kim.
\newblock The mit humanoid robot: Design, motion planning, and control for acrobatic behaviors.
\newblock In \emph{2020 IEEE-RAS 20th International Conference on Humanoid Robots (Humanoids)}, pages 1--8. IEEE, 2021.

\bibitem[Cui et~al.(2022)Cui, Wang, Shafiullah, and Pinto]{cui2022play}
Zichen~Jeff Cui, Yibin Wang, Nur Muhammad~Mahi Shafiullah, and Lerrel Pinto.
\newblock From play to policy: Conditional behavior generation from uncurated robot data.
\newblock \emph{arXiv preprint arXiv:2210.10047}, 2022.

\bibitem[Dallard et~al.(2023)Dallard, Benallegue, Kanehiro, and Kheddar]{dallard2023Sync}
Antonin Dallard, Mehdi Benallegue, Fumio Kanehiro, and Abderrahmane Kheddar.
\newblock Synchronized human-humanoid motion imitation.
\newblock \emph{IEEE Robotics and Automation Letters}, 8\penalty0 (7):\penalty0 4155--4162, 2023.
\newblock \doi{10.1109/LRA.2023.3280807}.

\bibitem[Darcet et~al.(2023)Darcet, Oquab, Mairal, and Bojanowski]{darcet2023vitneedreg}
Timothée Darcet, Maxime Oquab, Julien Mairal, and Piotr Bojanowski.
\newblock Vision transformers need registers, 2023.

\bibitem[Dariush et~al.(2008)Dariush, Gienger, Jian, Goerick, and Fujimura]{dariush2008whole}
Behzad Dariush, Michael Gienger, Bing Jian, Christian Goerick, and Kikuo Fujimura.
\newblock Whole body humanoid control from human motion descriptors.
\newblock In \emph{2008 IEEE International Conference on Robotics and Automation}, pages 2677--2684. IEEE, 2008.

\bibitem[Darvish et~al.(2019)Darvish, Tirupachuri, Romualdi, Rapetti, Ferigo, Chavez, and Pucci]{darvish2019wholebody}
Kourosh Darvish, Yeshasvi Tirupachuri, Giulio Romualdi, Lorenzo Rapetti, Diego Ferigo, Francisco Javier~Andrade Chavez, and Daniele Pucci.
\newblock Whole-body geometric retargeting for humanoid robots.
\newblock In \emph{2019 IEEE-RAS 19th International Conference on Humanoid Robots (Humanoids)}, pages 679--686, 2019.
\newblock \doi{10.1109/Humanoids43949.2019.9035059}.

\bibitem[Duan et~al.(2023)Duan, Pandit, Gadde, van Marum, Dao, Kim, and Fern]{duan2023learning}
Helei Duan, Bikram Pandit, Mohitvishnu~S Gadde, Bart~Jaap van Marum, Jeremy Dao, Chanho Kim, and Alan Fern.
\newblock Learning vision-based bipedal locomotion for challenging terrain.
\newblock \emph{arXiv preprint arXiv:2309.14594}, 2023.

\bibitem[Escontrela et~al.(2022)Escontrela, Peng, Yu, Zhang, Iscen, Goldberg, and Abbeel]{escontrela2022adversarial}
Alejandro Escontrela, Xue~Bin Peng, Wenhao Yu, Tingnan Zhang, Atil Iscen, Ken Goldberg, and Pieter Abbeel.
\newblock Adversarial motion priors make good substitutes for complex reward functions.
\newblock In \emph{2022 IEEE/RSJ International Conference on Intelligent Robots and Systems (IROS)}, pages 25--32. IEEE, 2022.

\bibitem[Fu et~al.(2022)Fu, Song, Wu, Yu, and Scaramuzza]{fu2022learning}
Jiawei Fu, Yunlong Song, Yan Wu, Fisher Yu, and Davide Scaramuzza.
\newblock Learning deep sensorimotor policies for vision-based autonomous drone racing, 2022.

\bibitem[Fu et~al.(2021)Fu, Kumar, Malik, and Pathak]{fu2021minimizing}
Zipeng Fu, Ashish Kumar, Jitendra Malik, and Deepak Pathak.
\newblock Minimizing energy consumption leads to the emergence of gaits in legged robots.
\newblock \emph{Conference on Robot Learning ({CoRL})}, 2021.

\bibitem[Fu et~al.(2024{\natexlab{a}})Fu, Zhao, Wu, Wetzstein, and Finn]{fu2024humanplus}
Zipeng Fu, Qingqing Zhao, Qi~Wu, Gordon Wetzstein, and Chelsea Finn.
\newblock Humanplus: Humanoid shadowing and imitation from humans, 2024{\natexlab{a}}.
\newblock URL \url{https://arxiv.org/abs/2406.10454}.

\bibitem[Fu et~al.(2024{\natexlab{b}})Fu, Zhao, and Finn]{fu2024mobile}
Zipeng Fu, Tony~Z Zhao, and Chelsea Finn.
\newblock Mobile aloha: Learning bimanual mobile manipulation with low-cost whole-body teleoperation.
\newblock \emph{arXiv preprint arXiv:2401.02117}, 2024{\natexlab{b}}.

\bibitem[Fuchioka et~al.(2023)Fuchioka, Xie, and Van~de Panne]{fuchioka2023opt}
Yuni Fuchioka, Zhaoming Xie, and Michiel Van~de Panne.
\newblock Opt-mimic: Imitation of optimized trajectories for dynamic quadruped behaviors.
\newblock In \emph{2023 IEEE International Conference on Robotics and Automation (ICRA)}, pages 5092--5098. IEEE, 2023.

\bibitem[Ha et~al.(2024)Ha, Gao, Fu, Tan, and Song]{ha2024umi}
Huy Ha, Yihuai Gao, Zipeng Fu, Jie Tan, and Shuran Song.
\newblock Umi on legs: Making manipulation policies mobile with manipulation-centric whole-body controllers.
\newblock \emph{arXiv preprint arXiv:2407.10353}, 2024.

\bibitem[He et~al.(2024{\natexlab{a}})He, Luo, He, Xiao, Zhang, Zhang, Kitani, Liu, and Shi]{he2024omnih2o}
Tairan He, Zhengyi Luo, Xialin He, Wenli Xiao, Chong Zhang, Weinan Zhang, Kris Kitani, Changliu Liu, and Guanya Shi.
\newblock Omnih2o: Universal and dexterous human-to-humanoid whole-body teleoperation and learning, 2024{\natexlab{a}}.
\newblock URL \url{https://arxiv.org/abs/2406.08858}.

\bibitem[He et~al.(2024{\natexlab{b}})He, Luo, Xiao, Zhang, Kitani, Liu, and Shi]{he2024h20}
Tairan He, Zhengyi Luo, Wenli Xiao, Chong Zhang, Kris Kitani, Changliu Liu, and Guanya Shi.
\newblock Learning human-to-humanoid real-time whole-body teleoperation.
\newblock \emph{arXiv preprint arXiv:2403.04436}, 2024{\natexlab{b}}.

\bibitem[He et~al.(2024{\natexlab{c}})He, Xiao, Lin, Luo, Xu, Jiang, Liu, Shi, Wang, Fan, and Zhu]{he2024hover}
Tairan He, Wenli Xiao, Toru Lin, Zhengyi Luo, Zhenjia Xu, Zhenyu Jiang, Changliu Liu, Guanya Shi, Xiaolong Wang, Linxi Fan, and Yuke Zhu.
\newblock Hover: Versatile neural whole-body controller for humanoid robots.
\newblock \emph{arXiv preprint arXiv:2410.21229}, 2024{\natexlab{c}}.

\bibitem[Hirai et~al.(1998)Hirai, Hirose, Haikawa, and Takenaka]{hirai1998development}
Kazuo Hirai, Masato Hirose, Yuji Haikawa, and Toru Takenaka.
\newblock The development of honda humanoid robot.
\newblock In \emph{Proceedings. 1998 IEEE international conference on robotics and automation (Cat. No. 98CH36146)}, volume~2, pages 1321--1326. IEEE, 1998.

\bibitem[Hutter et~al.(2016)Hutter, Gehring, Jud, Lauber, Bellicoso, Tsounis, Hwangbo, Bodie, Fankhauser, Bloesch, et~al.]{hutter2016anymal}
Marco Hutter, Christian Gehring, Dominic Jud, Andreas Lauber, C~Dario Bellicoso, Vassilios Tsounis, Jemin Hwangbo, Karen Bodie, Peter Fankhauser, Michael Bloesch, et~al.
\newblock Anymal-a highly mobile and dynamic quadrupedal robot.
\newblock In \emph{IROS}, 2016.

\bibitem[Ji et~al.(2024)Ji, Peng, Liu, Li, Yang, Cheng, and Wang]{ji2024exbody2}
Mazeyu Ji, Xuanbin Peng, Fangchen Liu, Jialong Li, Ge~Yang, Xuxin Cheng, and Xiaolong Wang.
\newblock Exbody2: Advanced expressive humanoid whole-body control.
\newblock \emph{arXiv preprint arXiv:2412.13196}, 2024.

\bibitem[Kajita(2001)]{kajita2001simple}
Shuuji Kajita.
\newblock A simple modeling for a biped walking pattern generation.
\newblock In \emph{Proceedings of the International Conference on Intelligent Robotics and Systems, Maui, HI, USA}, volume~29, 2001.

\bibitem[Kato(1973)]{kato1973development}
Ichiro Kato.
\newblock Development of wabot 1.
\newblock \emph{Biomechanism}, 2:\penalty0 173--214, 1973.

\bibitem[Khazatsky et~al.(2024)Khazatsky, Pertsch, Nair, Balakrishna, Dasari, Karamcheti, Nasiriany, Srirama, Chen, Ellis, Fagan, Hejna, Itkina, Lepert, Ma, Miller, Wu, Belkhale, Dass, Ha, Jain, Lee, Lee, Memmel, Park, Radosavovic, Wang, Zhan, Black, Chi, Hatch, Lin, Lu, Mercat, Rehman, Sanketi, Sharma, Simpson, Vuong, Walke, Wulfe, Xiao, Yang, Yavary, Zhao, Agia, Baijal, Castro, Chen, Chen, Chung, Drake, Foster, Gao, Herrera, Heo, Hsu, Hu, Jackson, Le, Li, Lin, Lin, Ma, Maddukuri, Mirchandani, Morton, Nguyen, O'Neill, Scalise, Seale, Son, Tian, Tran, Wang, Wu, Xie, Yang, Yin, Zhang, Bastani, Berseth, Bohg, Goldberg, Gupta, Gupta, Jayaraman, Lim, Malik, Martín-Martín, Ramamoorthy, Sadigh, Song, Wu, Yip, Zhu, Kollar, Levine, and Finn]{khazatsky2024droid}
Alexander Khazatsky, Karl Pertsch, Suraj Nair, Ashwin Balakrishna, Sudeep Dasari, Siddharth Karamcheti, Soroush Nasiriany, Mohan~Kumar Srirama, Lawrence~Yunliang Chen, Kirsty Ellis, Peter~David Fagan, Joey Hejna, Masha Itkina, Marion Lepert, Yecheng~Jason Ma, Patrick~Tree Miller, Jimmy Wu, Suneel Belkhale, Shivin Dass, Huy Ha, Arhan Jain, Abraham Lee, Youngwoon Lee, Marius Memmel, Sungjae Park, Ilija Radosavovic, Kaiyuan Wang, Albert Zhan, Kevin Black, Cheng Chi, Kyle~Beltran Hatch, Shan Lin, Jingpei Lu, Jean Mercat, Abdul Rehman, Pannag~R Sanketi, Archit Sharma, Cody Simpson, Quan Vuong, Homer~Rich Walke, Blake Wulfe, Ted Xiao, Jonathan~Heewon Yang, Arefeh Yavary, Tony~Z. Zhao, Christopher Agia, Rohan Baijal, Mateo~Guaman Castro, Daphne Chen, Qiuyu Chen, Trinity Chung, Jaimyn Drake, Ethan~Paul Foster, Jensen Gao, David~Antonio Herrera, Minho Heo, Kyle Hsu, Jiaheng Hu, Donovon Jackson, Charlotte Le, Yunshuang Li, Kevin Lin, Roy Lin, Zehan Ma, Abhiram Maddukuri, Suvir Mirchandani, Daniel Morton, Tony Nguyen,
  Abigail O'Neill, Rosario Scalise, Derick Seale, Victor Son, Stephen Tian, Emi Tran, Andrew~E. Wang, Yilin Wu, Annie Xie, Jingyun Yang, Patrick Yin, Yunchu Zhang, Osbert Bastani, Glen Berseth, Jeannette Bohg, Ken Goldberg, Abhinav Gupta, Abhishek Gupta, Dinesh Jayaraman, Joseph~J Lim, Jitendra Malik, Roberto Martín-Martín, Subramanian Ramamoorthy, Dorsa Sadigh, Shuran Song, Jiajun Wu, Michael~C. Yip, Yuke Zhu, Thomas Kollar, Sergey Levine, and Chelsea Finn.
\newblock Droid: A large-scale in-the-wild robot manipulation dataset.
\newblock 2024.

\bibitem[Kumar et~al.(2021)Kumar, Fu, Pathak, and Malik]{kumar2021rma}
Ashish Kumar, Zipeng Fu, Deepak Pathak, and Jitendra Malik.
\newblock Rma: Rapid motor adaptation for legged robots.
\newblock \emph{arXiv preprint arXiv:2107.04034}, 2021.

\bibitem[Li et~al.(2021)Li, Cheng, Peng, Abbeel, Levine, Berseth, and Sreenath]{li2021reinforcement}
Zhongyu Li, Xuxin Cheng, Xue~Bin Peng, Pieter Abbeel, Sergey Levine, Glen Berseth, and Koushil Sreenath.
\newblock Reinforcement learning for robust parameterized locomotion control of bipedal robots.
\newblock In \emph{2021 IEEE International Conference on Robotics and Automation (ICRA)}, pages 2811--2817. IEEE, 2021.

\bibitem[Li et~al.(2024)Li, Peng, Abbeel, Levine, Berseth, and Sreenath]{li2024reinforcement}
Zhongyu Li, Xue~Bin Peng, Pieter Abbeel, Sergey Levine, Glen Berseth, and Koushil Sreenath.
\newblock Reinforcement learning for versatile, dynamic, and robust bipedal locomotion control.
\newblock \emph{arXiv preprint arXiv:2401.16889}, 2024.

\bibitem[Liao et~al.(2024)Liao, Zhang, Huang, Huang, Li, and Sreenath]{liao2024berkeley}
Qiayuan Liao, Bike Zhang, Xuanyu Huang, Xiaoyu Huang, Zhongyu Li, and Koushil Sreenath.
\newblock Berkeley humanoid: A research platform for learning-based control.
\newblock \emph{arXiv preprint arXiv:2407.21781}, 2024.

\bibitem[Liu et~al.(2024)Liu, Gu, Cai, Zhou, Zhao, Jung, Ha, Chen, Xu, and Zhao]{liu2024opt2skill}
Fukang Liu, Zhaoyuan Gu, Yilin Cai, Ziyi Zhou, Shijie Zhao, Hyunyoung Jung, Sehoon Ha, Yue Chen, Danfei Xu, and Ye~Zhao.
\newblock Opt2skill: Imitating dynamically-feasible whole-body trajectories for versatile humanoid loco-manipulation.
\newblock \emph{arXiv preprint arXiv:2409.20514}, 2024.

\bibitem[Lu et~al.(2024)Lu, Cheng, Li, Yang, Ji, Yuan, Yang, Yi, and Wang]{lu2024pmp}
Chenhao Lu, Xuxin Cheng, Jialong Li, Shiqi Yang, Mazeyu Ji, Chengjing Yuan, Ge~Yang, Sha Yi, and Xiaolong Wang.
\newblock Mobile-television: Predictive motion priors for humanoid whole-body control.
\newblock \emph{arXiv preprint arXiv:2412.07773}, 2024.

\bibitem[Mahmood et~al.(2019)Mahmood, Ghorbani, F.~Troje, Pons-Moll, and Black]{AMASS:2019}
Naureen Mahmood, Nima Ghorbani, Nikolaus F.~Troje, Gerard Pons-Moll, and Michael~J. Black.
\newblock Amass: Archive of motion capture as surface shapes.
\newblock In \emph{The IEEE International Conference on Computer Vision (ICCV)}, Oct 2019.
\newblock URL \url{https://amass.is.tue.mpg.de}.

\bibitem[Makoviychuk et~al.(2021)Makoviychuk, Wawrzyniak, Guo, Lu, Storey, Macklin, Hoeller, Rudin, Allshire, Handa, et~al.]{makoviychuk2021isaac}
Viktor Makoviychuk, Lukasz Wawrzyniak, Yunrong Guo, Michelle Lu, Kier Storey, Miles Macklin, David Hoeller, Nikita Rudin, Arthur Allshire, Ankur Handa, et~al.
\newblock Isaac gym: High performance gpu-based physics simulation for robot learning.
\newblock \emph{arXiv preprint arXiv:2108.10470}, 2021.

\bibitem[Margolis and Agrawal(2022)]{margolis2022walktheseways}
Gabriel~B Margolis and Pulkit Agrawal.
\newblock Walk these ways: Tuning robot control for generalization with multiplicity of behavior.
\newblock \emph{Conference on Robot Learning}, 2022.

\bibitem[Margolis et~al.(2021)Margolis, Chen, Paigwar, Fu, Kim, Kim, and Agrawal]{margolis2021learning}
Gabriel~B Margolis, Tao Chen, Kartik Paigwar, Xiang Fu, Donghyun Kim, Sangbae Kim, and Pulkit Agrawal.
\newblock Learning to jump from pixels.
\newblock \emph{arXiv preprint arXiv:2110.15344}, 2021.

\bibitem[Margolis et~al.(2022)Margolis, Yang, Paigwar, Chen, and Agrawal]{margolis2022rapid}
Gabriel~B Margolis, Ge~Yang, Kartik Paigwar, Tao Chen, and Pulkit Agrawal.
\newblock Rapid locomotion via reinforcement learning.
\newblock \emph{arXiv preprint arXiv:2205.02824}, 2022.

\bibitem[Mastalli et~al.(2020)Mastalli, Budhiraja, Merkt, Saurel, Hammoud, Naveau, Carpentier, Righetti, Vijayakumar, and Mansard]{mastalli20crocoddyl}
Carlos Mastalli, Rohan Budhiraja, Wolfgang Merkt, Guilhem Saurel, Bilal Hammoud, Maximilien Naveau, Justin Carpentier, Ludovic Righetti, Sethu Vijayakumar, and Nicolas Mansard.
\newblock {Crocoddyl: An Efficient and Versatile Framework for Multi-Contact Optimal Control}.
\newblock In \emph{IEEE International Conference on Robotics and Automation (ICRA)}, 2020.

\bibitem[Mastalli et~al.(2022)Mastalli, Merkt, Marti-Saumell, Ferrolho, Sol{\`a}, Mansard, and Vijayakumar]{mastalli2022fddp}
Carlos Mastalli, Wolfgang Merkt, Josep Marti-Saumell, Henrique Ferrolho, Joan Sol{\`a}, Nicolas Mansard, and Sethu Vijayakumar.
\newblock A feasibility-driven approach to control-limited ddp.
\newblock \emph{Autonomous Robots}, 2022.

\bibitem[Mendonca et~al.(2023)Mendonca, Bahl, and Pathak]{mendonca2023alan}
Russell Mendonca, Shikhar Bahl, and Deepak Pathak.
\newblock Alan : Autonomously exploring robotic agents in the real world.
\newblock \emph{ICRA}, 2023.

\bibitem[Miura and Shimoyama(1984)]{miura1984dynamic}
Hirofumi Miura and Isao Shimoyama.
\newblock Dynamic walk of a biped.
\newblock \emph{IJRR}, 1984.

\bibitem[Moro and Sentis(2019)]{moro2019whole}
Federico~L Moro and Luis Sentis.
\newblock Whole-body control of humanoid robots.
\newblock \emph{Humanoid Robotics: A reference, Springer, Dordrecht}, 2019.

\bibitem[Oquab et~al.(2023)Oquab, Darcet, Moutakanni, Vo, Szafraniec, Khalidov, Fernandez, Haziza, Massa, El-Nouby, Howes, Huang, Xu, Sharma, Li, Galuba, Rabbat, Assran, Ballas, Synnaeve, Misra, Jegou, Mairal, Labatut, Joulin, and Bojanowski]{oquab2023dinov2}
Maxime Oquab, Timothée Darcet, Theo Moutakanni, Huy~V. Vo, Marc Szafraniec, Vasil Khalidov, Pierre Fernandez, Daniel Haziza, Francisco Massa, Alaaeldin El-Nouby, Russell Howes, Po-Yao Huang, Hu~Xu, Vasu Sharma, Shang-Wen Li, Wojciech Galuba, Mike Rabbat, Mido Assran, Nicolas Ballas, Gabriel Synnaeve, Ishan Misra, Herve Jegou, Julien Mairal, Patrick Labatut, Armand Joulin, and Piotr Bojanowski.
\newblock Dinov2: Learning robust visual features without supervision, 2023.

\bibitem[Padalkar et~al.(2023)Padalkar, Pooley, Jain, Bewley, Herzog, Irpan, Khazatsky, Rai, Singh, Brohan, et~al.]{padalkar2023open}
Abhishek Padalkar, Acorn Pooley, Ajinkya Jain, Alex Bewley, Alex Herzog, Alex Irpan, Alexander Khazatsky, Anant Rai, Anikait Singh, Anthony Brohan, et~al.
\newblock Open x-embodiment: Robotic learning datasets and rt-x models.
\newblock \emph{arXiv preprint arXiv:2310.08864}, 2023.

\bibitem[Pari et~al.(2021)Pari, Shafiullah, Arunachalam, and Pinto]{pari2021surprising}
Jyothish Pari, Nur~Muhammad Shafiullah, Sridhar~Pandian Arunachalam, and Lerrel Pinto.
\newblock The surprising effectiveness of representation learning for visual imitation.
\newblock \emph{arXiv preprint arXiv:2112.01511}, 2021.

\bibitem[Penco et~al.(2019)Penco, Scianca, Modugno, Lanari, Oriolo, and Ivaldi]{penco2019multimode}
Luigi Penco, Nicola Scianca, Valerio Modugno, Leonardo Lanari, Giuseppe Oriolo, and Serena Ivaldi.
\newblock A multimode teleoperation framework for humanoid loco-manipulation: An application for the icub robot.
\newblock \emph{IEEE Robotics and Automation Magazine}, 26\penalty0 (4):\penalty0 73--82, 2019.
\newblock \doi{10.1109/MRA.2019.2941245}.

\bibitem[Peng et~al.(2020)Peng, Coumans, Zhang, Lee, Tan, and Levine]{Peng2020-ty}
Xue~Bin Peng, Erwin Coumans, Tingnan Zhang, Tsang-Wei Lee, Jie Tan, and Sergey Levine.
\newblock Learning agile robotic locomotion skills by imitating animals.
\newblock April 2020.

\bibitem[Peng et~al.(2021{\natexlab{a}})Peng, Ma, Abbeel, Levine, and Kanazawa]{Peng2021-sq}
Xue~Bin Peng, Ze~Ma, Pieter Abbeel, Sergey Levine, and Angjoo Kanazawa.
\newblock {AMP}: adversarial motion priors for stylized physics-based character control.
\newblock \emph{ACM Trans. Graph.}, 40\penalty0 (4):\penalty0 1--20, July 2021{\natexlab{a}}.

\bibitem[Peng et~al.(2021{\natexlab{b}})Peng, Ma, Abbeel, Levine, and Kanazawa]{peng2021amp}
Xue~Bin Peng, Ze~Ma, Pieter Abbeel, Sergey Levine, and Angjoo Kanazawa.
\newblock Amp: Adversarial motion priors for stylized physics-based character control.
\newblock \emph{ACM Transactions on Graphics (ToG)}, 40\penalty0 (4):\penalty0 1--20, 2021{\natexlab{b}}.

\bibitem[Qin et~al.(2023)Qin, Yang, Huang, Van~Wyk, Su, Wang, Chao, and Fox]{qin2023anyteleop}
Yuzhe Qin, Wei Yang, Binghao Huang, Karl Van~Wyk, Hao Su, Xiaolong Wang, Yu-Wei Chao, and Dieter Fox.
\newblock Anyteleop: A general vision-based dexterous robot arm-hand teleoperation system.
\newblock In \emph{Robotics: Science and Systems}, 2023.

\bibitem[Radosavovic et~al.(2023)Radosavovic, Xiao, Zhang, Darrell, Malik, and Sreenath]{radosavovic2023learning}
Ilija Radosavovic, Tete Xiao, Bike Zhang, Trevor Darrell, Jitendra Malik, and Koushil Sreenath.
\newblock Learning humanoid locomotion with transformers.
\newblock \emph{arXiv preprint arXiv:2303.03381}, 2023.

\bibitem[Raibert(1986)]{raibert1986legged}
Marc~H Raibert.
\newblock \emph{Legged robots that balance}.
\newblock MIT press, 1986.

\bibitem[Schulman et~al.(2017)Schulman, Wolski, Dhariwal, Radford, and Klimov]{schulman2017proximal}
John Schulman, Filip Wolski, Prafulla Dhariwal, Alec Radford, and Oleg Klimov.
\newblock Proximal policy optimization algorithms.
\newblock \emph{arXiv preprint arXiv:1707.06347}, 2017.

\bibitem[Seo et~al.(2023)Seo, Han, Sim, Bang, Gonzalez, Sentis, and Zhu]{seo2023deep}
Mingyo Seo, Steve Han, Kyutae Sim, Seung~Hyeon Bang, Carlos Gonzalez, Luis Sentis, and Yuke Zhu.
\newblock Deep imitation learning for humanoid loco-manipulation through human teleoperation.
\newblock In \emph{2023 IEEE-RAS 22nd International Conference on Humanoid Robots (Humanoids)}, pages 1--8. IEEE, 2023.

\bibitem[Shi et~al.(2024)Shi, Hu, Zhao, Sharma, Pertsch, Luo, Levine, and Finn]{shi2024yell}
Lucy~Xiaoyang Shi, Zheyuan Hu, Tony~Z. Zhao, Archit Sharma, Karl Pertsch, Jianlan Luo, Sergey Levine, and Chelsea Finn.
\newblock Yell at your robot: Improving on-the-fly from language corrections.
\newblock \emph{arXiv preprint arXiv: 2403.12910}, 2024.

\bibitem[Shridhar et~al.(2023)Shridhar, Manuelli, and Fox]{shridhar2023perceiver}
Mohit Shridhar, Lucas Manuelli, and Dieter Fox.
\newblock Perceiver-actor: A multi-task transformer for robotic manipulation.
\newblock In \emph{Conference on Robot Learning}, pages 785--799. PMLR, 2023.

\bibitem[Siekmann et~al.(2021)Siekmann, Green, Warila, Fern, and Hurst]{siekmann2021blind}
Jonah Siekmann, Kevin Green, John Warila, Alan Fern, and Jonathan Hurst.
\newblock Blind bipedal stair traversal via sim-to-real reinforcement learning.
\newblock \emph{arXiv preprint arXiv:2105.08328}, 2021.

\bibitem[Sugihara(2011)]{Sugihara2011}
Tomomichi Sugihara.
\newblock Solvability-unconcerned inverse kinematics by the levenberg–marquardt method.
\newblock \emph{IEEE Transactions on Robotics}, 27\penalty0 (5):\penalty0 984--991, 2011.
\newblock \doi{10.1109/TRO.2011.2148230}.
\newblock URL \url{https://ieeexplore.ieee.org/document/5784347}.

\bibitem[Tessler et~al.(2023)Tessler, Kasten, Guo, Mannor, Chechik, and Peng]{tessler2023calm}
Chen Tessler, Yoni Kasten, Yunrong Guo, Shie Mannor, Gal Chechik, and Xue~Bin Peng.
\newblock Calm: Conditional adversarial latent models for directable virtual characters.
\newblock In \emph{ACM SIGGRAPH 2023 Conference Proceedings}, SIGGRAPH '23, New York, NY, USA, 2023. Association for Computing Machinery.
\newblock ISBN 9798400701597.
\newblock \doi{10.1145/3588432.3591541}.
\newblock URL \url{https://doi.org/10.1145/3588432.3591541}.

\bibitem[Wang et~al.(2023)Wang, Fan, Sun, Zhang, Fei-Fei, Xu, Zhu, and Anandkumar]{wang2023mimicplay}
Chen Wang, Linxi Fan, Jiankai Sun, Ruohan Zhang, Li~Fei-Fei, Danfei Xu, Yuke Zhu, and Anima Anandkumar.
\newblock Mimicplay: Long-horizon imitation learning by watching human play.
\newblock \emph{arXiv preprint arXiv:2302.12422}, 2023.

\bibitem[Wang et~al.(2020)Wang, Guo, Shugrina, and Fidler]{wang2020unicon}
Tingwu Wang, Yunrong Guo, Maria Shugrina, and Sanja Fidler.
\newblock Unicon: Universal neural controller for physics-based character motion, 2020.

\bibitem[Westervelt et~al.(2003)Westervelt, Grizzle, and Koditschek]{westervelt2003hybrid}
Eric~R Westervelt, Jessy~W Grizzle, and Daniel~E Koditschek.
\newblock Hybrid zero dynamics of planar biped walkers.
\newblock \emph{IEEE transactions on automatic control}, 48\penalty0 (1):\penalty0 42--56, 2003.

\bibitem[Yang et~al.(2023{\natexlab{a}})Yang, Chen, Ma, Zheng, Chen, Nguyen, and Wang]{yang2023generalized}
Ruihan Yang, Zhuoqun Chen, Jianhan Ma, Chongyi Zheng, Yiyu Chen, Quan Nguyen, and Xiaolong Wang.
\newblock Generalized animal imitator: Agile locomotion with versatile motion prior.
\newblock \emph{arXiv preprint arXiv:2310.01408}, 2023{\natexlab{a}}.

\bibitem[Yang et~al.(2023{\natexlab{b}})Yang, Yang, and Wang]{yang2023neural}
Ruihan Yang, Ge~Yang, and Xiaolong Wang.
\newblock Neural volumetric memory for visual locomotion control.
\newblock In \emph{Proceedings of the IEEE/CVF Conference on Computer Vision and Pattern Recognition}, pages 1430--1440, 2023{\natexlab{b}}.

\bibitem[Yin et~al.(2007)Yin, Loken, and Van~de Panne]{yin2007simbicon}
KangKang Yin, Kevin Loken, and Michiel Van~de Panne.
\newblock Simbicon: Simple biped locomotion control.
\newblock \emph{ACM Transactions on Graphics}, 2007.

\bibitem[Ze et~al.(2024)Ze, Zhang, Zhang, Hu, Wang, and Xu]{Ze2024DP3}
Yanjie Ze, Gu~Zhang, Kangning Zhang, Chenyuan Hu, Muhan Wang, and Huazhe Xu.
\newblock 3d diffusion policy: Generalizable visuomotor policy learning via simple 3d representations.
\newblock In \emph{Proceedings of Robotics: Science and Systems (RSS)}, 2024.

\bibitem[Zhang et~al.(2024)Zhang, Cui, Yan, Sun, Duan, Zhang, and Xu]{zhang2024whole}
Qiang Zhang, Peter Cui, David Yan, Jingkai Sun, Yiqun Duan, Arthur Zhang, and Renjing Xu.
\newblock Whole-body humanoid robot locomotion with human reference.
\newblock \emph{arXiv preprint arXiv:2402.18294}, 2024.

\bibitem[Zhao et~al.(2023)Zhao, Kumar, Levine, and Finn]{zhao2023learning}
Tony~Z Zhao, Vikash Kumar, Sergey Levine, and Chelsea Finn.
\newblock Learning fine-grained bimanual manipulation with low-cost hardware.
\newblock \emph{arXiv preprint arXiv:2304.13705}, 2023.

\bibitem[Zhuang et~al.(2023)Zhuang, Fu, Wang, Atkeson, Schwertfeger, Finn, and Zhao]{zhuang2023robot}
Ziwen Zhuang, Zipeng Fu, Jianren Wang, Christopher Atkeson, Sören Schwertfeger, Chelsea Finn, and Hang Zhao.
\newblock Robot parkour learning.
\newblock In \emph{Conference on Robot Learning ({CoRL})}, 2023.

\end{thebibliography}

\clearpage
\appendices
\setcounter{section}{0}
\renewcommand{\thesection}{\Alph{section}}
\section{Retargeting}
\label{supp:retargeting}
The hand retargeting problem is formulated as the following objective function:
\begin{align*}
\min_{\mathbf{q}_\mathrm{hand}} & \quad \sum_{i=1}^n \| \mathbf{p}_i^\mathrm{human} - \mathbf{p}_i^\mathrm{robot} \|^2 \\
\mathbf{p}_i^\mathrm{robot} &= FK(\mathbf{q}_\mathrm{hand}) \\
\text{subject to} & \quad \mathbf{q}_\mathrm{min} \leq \mathbf{q}_\mathrm{hand} \leq \mathbf{q}_\mathrm{max}
\end{align*}
This objective aims at minimizing the distances between corresponding keypoint vectors. Here, $\mathbf{p}_i^\mathrm{human}$ denotes the $i^\mathrm{th}$ (scaled) keypoint vector of the human hand, while $\mathbf{p}_i^\mathrm{robot}$ represents the corresponding vector of the robot hand. $\mathbf{p}_i^\mathrm{robot}= FK(\mathbf{q}_\mathrm{hand})$ is derived using forward kinematics from robot hand joint angles $\mathbf{q}_\mathrm{hand}$. Since our system features a three-fingered robot hand,five vectors are employed for retargeting: three vectors ranging from the wrist to each fingertip and two vectors ranging from the thumb tip to the index and middle fingertips. Vectors originating from the wrist ensure that the overall robot hand pose closely resembles the human hand's pose, while vectors from the thumb tip facilitate precise and fine manipulations. 

\section{RL Training Details}
\label{supp:RL}
\textbf{Teacher Training Curriculum} We use a carefully designed curriculum to regulate different tasks during RL training. Initially, the lower policy is treated as a pure locomotion policy with randomized target height. Then, we add delay to reduce sim-to-real gap. We then randomize target torso roll, pitch and yaw and activate related rewards. We finally sample arm actions from AMASS dataset and directly set the arm joint target position in the environment. Following the training schedule, the teacher policy gradually master complex whole-body control setting and be able to supervise a real world student.

\begin{table}[h]
    \centering
    \begin{tabular*}{\linewidth}{@{\extracolsep{\fill}} l c c}
        \toprule
        \textbf{Curriculum} & \textbf{Begin Global Step} & \textbf{Randomize Range} \\
        \midrule
        Target Height & 100 & 0.5 $\sim$ 0.8 \\
        Target Torso Roll & 2000 & -0.7 $\sim$ 0.7 \\
        Target Torso Pitch & 2000 & -0.52 $\sim$ 1.57 \\
        Target Torso Yaw & 2000 & -1.57 $\sim$ 1.57 \\
        Target Arm & 4000 & AMASS \\
        \bottomrule
    \end{tabular*}
    \caption{Curriculum Random Sample Steps}
    \label{tab:random_curriculum}
\end{table}

Table ~\ref{tab:random_curriculum} shows the detail of our randomized curriculum schedule. When the global step is less than the curriculum begin global step, the target height is set to $0.8$, the target torso roll, pitch, and yaw are set to $0$, the target arm joint positions are set to the default arm joint positions. When the global step reaches the curriculum begin global step, target height, target torso roll, pitch, and yaw are randomly sampled in the ranges specified in the table, and the target arm joint positions are sampled from the AMASS dataset.

\begin{table}[h]
    \centering
    \begin{tabular*}{\linewidth}{@{\extracolsep{\fill}} l c c c}
        \toprule
        \textbf{Curriculum} & \textbf{Begin Global Step} & \textbf{Final Global Step} & \textbf{Range} \\
        \midrule
        Gait frequency & 5000 & 10000 & 1.3 $\sim$ 1.0 \\
        Stance Rate & 1000 & 2000 & 0.2 $\sim$ 0.5 \\
        \bottomrule
    \end{tabular*}
    \caption{Curriculum Linear Sample Steps}
    \label{tab:linear_curriculum}
\end{table}

Table ~\ref{tab:linear_curriculum} shows the detail of our linear sampled curriculum schedule.
The gait frequency can be formulated as follows:
The Gait Frequency and Stance Rate are defined as piecewise functions based on the Global Step:

\[
\text{Gait Frequency}(s) =
\begin{cases}
1.3 & s < 5k \\
1.3 - \frac{0.3}{10k - 5k} (s - 5k), & 5k \leq s \leq 10k \\
1.0 & s > 10k
\end{cases}
\]

\[
\text{Stance Rate}(s) =
\begin{cases} 
0.2 & s < 5k \\
0.2 + \frac{0.3}{2k - 1k} (s - 1k), & 1k \leq s \leq 2k \\
0.5, & s > 2k
\end{cases}
\]

where \( s \) represents the global step.

\section{Deployment Details}
\label{supp:deployment}

\begin{figure}[t]
  \centering
  \includegraphics[width=0.8\linewidth]{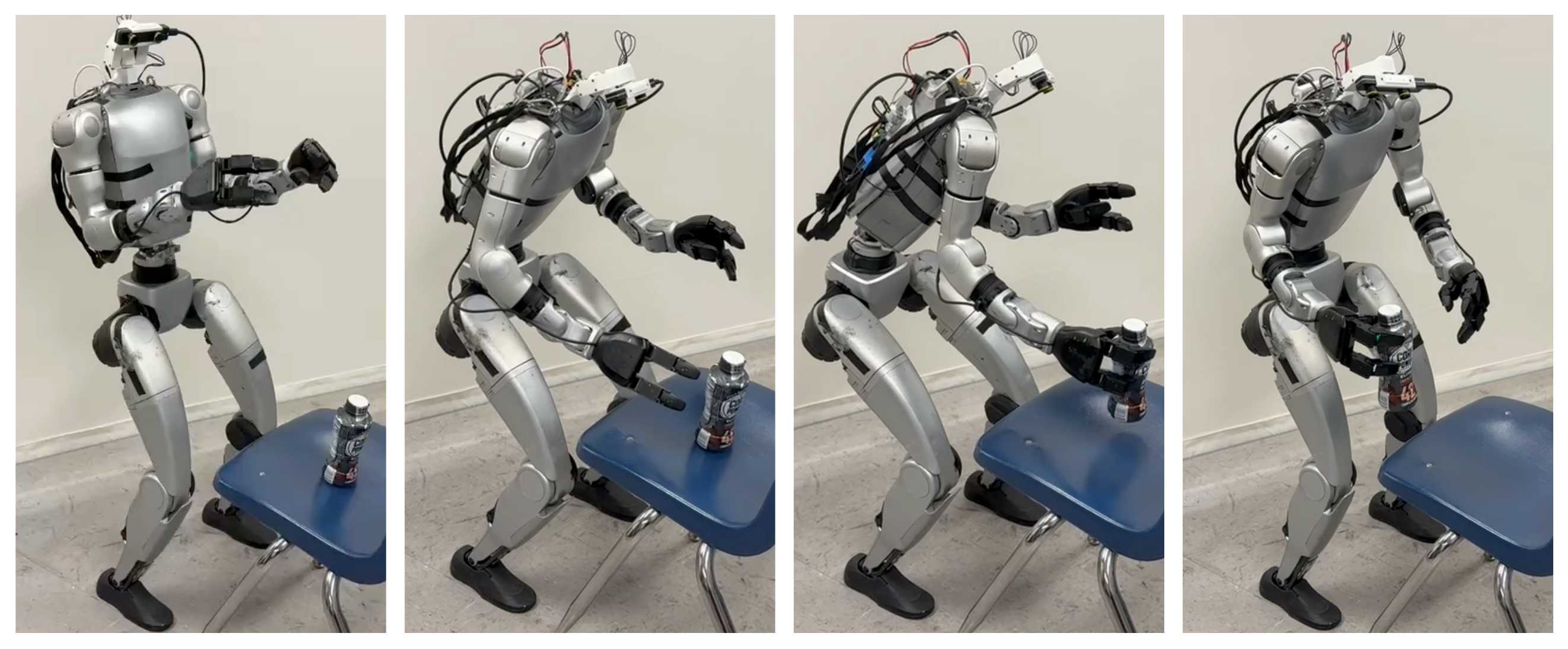}
  \caption{Fully onboard whole-body teleoperation. \label{fig:onboard}}
  \vspace{-0.2in}
\end{figure}

Some of our experiments are conducted with an external PC for easier debugging with an RTX 4090 GPU. Nonetheless, our entire teleoperation system and RL policy can be deployed onboard a Jetson Orin NX with 50Hz inference frequency, as shown in Fig.~\ref{fig:onboard}. We use OpenTelevision\cite{cheng2024tv} to stream images and human poses between VR devices and the robot. 

\end{document}